\documentclass[twoside]{article}
\usepackage{ecj,palatino,epsfig,latexsym,natbib}
\usepackage{graphicx}
\usepackage{amssymb,amsfonts}
\usepackage{textcomp}
\usepackage{xcolor}

\usepackage[]{algpseudocode}
\usepackage[colorlinks = true,
            linkcolor = red,
            urlcolor  = cyan,
            citecolor = blue,
            anchorcolor = blue,bookmarks=false]{hyperref}
\usepackage{subfig}    
\usepackage{float}
\usepackage{tikz}
\usepackage{multirow}
\usepackage{colortbl}
\usepackage{arydshln}
\usepackage{amsmath}
\usepackage{algorithm}
\usepackage{array}
\algnewcommand\algorithmicforeach{\textbf{for each}}
\algdef{S}[FOR]{ForEach}[1]{\algorithmicforeach\ #1\ \algorithmicdo}

\begin{document}

\ecjHeader{x}{x}{xxx-xxx}{201X}{Faster Convergence in MOEA/D}{Lavinas, Y., Aranha,  C., Ladeira, M.}
\title{\bf Faster Convergence in Multi-Objective Optimization Algorithms Based on Decomposition}  

\author{\name{\bf Yuri Lavinas} \hfill \addr{lavinas.yuri.xp@alumni.tsukuba.ac.jp}\\ 
        \addr{School of Systems and Information Engineering, University of Tsukuba, Japan}%Department of Science, My University, MyTown, Zip, Country
    \AND
       \name{\bf Marcelo Ladeira} \hfill \addr{mladeira@unb.br}\\
        \addr{Department of Computer Science, University of Brasilia, Brazil}
    \AND    
       \name{\bf Claus Aranha} \hfill \addr{caranha@cs.tsukuba.ac.jp}\\
        \addr{School of Systems and Information Engineering, University of Tsukuba, Japan}
}

\maketitle

\begin{abstract}
    The Resource Allocation approach (RA) improves the performance of MOEA/D by maintaining a big population and updating few solutions each generation. However, most of the studies on RA generally focused on the properties of different Resource Allocation metrics. Thus, it is still uncertain what the main factors are that lead to increments in performance of MOEA/D with RA. This study investigates the effects of MOEA/D with the Partial Update Strategy in an extensive set of MOPs to generate insights into correspondences of MOEA/D with the Partial Update and MOEA/D with small population size and big population size. Our work undertakes an in-depth analysis of the populational dynamics behaviour considering their final approximation Pareto sets, anytime hypervolume performance, attained regions and number of unique non-dominated solutions. Our results indicate that MOEA/D with Partial Update progresses with the search as fast as MOEA/D with small population size and explores the search space as MOEA/D with big population size. MOEA/D with Partial Update can mitigate common problems related to population size choice with better convergence speed in most MOPs, as shown by the results of hypervolume and number of unique non-dominated solutions, the anytime performance and Empirical Attainment Function indicates.
\end{abstract}


\section{Introduction}
\label{section:intro}

Multi-objective Optimisation Problems (MOPs) are problems with two or more conflicting objective functions that are optimised simultaneously. The Multi-Objective Evolutionary
Algorithm Based on Decomposition (MOEA/D) is a popular
and efficient algorithm for finding good sets of trade-off
solutions for MOPs (\cite{zhang2007moea}). The key idea of MOEA/D is to create a linear decomposition of the multi-objective optimisation problem into a set of single-objective sub-problems. These sub-problems are solved in parallel. Each individual of the population is assigned to be the "incumbent" solution of a sub-problem. In the original MOEA/D, all sub-problems are considered to be equally important at any time in the optimisation process. However, more recent research has indicated that the MOEA/D may waste computational effort by improving specific incumbent solutions that are not very promising (\cite{bezerra2015comparing}). Treating all sub-problems with the same importance can be a critical issue, especially in problems where the computation cost of the fitness function is high and the evaluation budget is low (\cite{kohira2018proposal}).
%and some of its earlier do not discriminate between sub-problems. However, it has become clear that focusing computational effort on specific subsets of these sub-problems can substantially improve the algorithm's performance. Moreover, MOEA/D may waste computational effort by trying to improve solutions that are not very promising~\cite{bezerra2015comparing}. This waste of computational effort can be a critical issue,  specifically in MOPs which require costly simulations to evaluate solutions~\cite{kohira2018proposal}. 

To address this issue, several works have investigated methods to allocate different amounts of computational effort to sub-problems (\cite{zhang2009performance, zhou2016all, lavinas2019improving, lavinas2019using, wangNewResourceAllocation2019,pruvost2020subproblemselection, lavinas2020moea}). 
The general approach is to use different indicators (priority functions) to determine how many fitness evaluations are allocated to each sub-problem. 
At each iteration of the algorithm, only a subset of the population is updated and evaluated, selected according to the priority function used. 
We call the selected individuals on each iteration the \emph{working population}.
These approaches are collectively known as Resource Allocation (RA) and have resulted in consistent performance improvements for the MOEA/D.

% , and the non-selected ones the \emph{inert population}.

Existing RA research placed considerable importance on the properties of different priority functions. However, recent studies from \cite{pruvost2020subproblemselection}, and \cite{lavinas2020moea} have found that the improvements in performance obtained from RA could be attributed to the reduction in the size of the working population. These results raise the question: are the improvements from RA due only to the reduced population size, or are there other factors also at work in the improvement observed in the various RA approaches? There is no research so far studying the population dynamics of MOEA/D with RAs.

To fill in this gap, we investigate and quantify the effects of RA on MOEA/D against MOEA/D with different population sizes on continuous problems. To achieve this, we study the correspondences of MOEA/D using RA against (1) no RA on small population size and (2) no RA on larger population size. We perform an experimental investigation and systematically analyse the impact of the proportion parameter $n$ on the performance of the MOEA/D on 70 different MOPs. This experimental work presented here provides one of the first investigations into how MOEA/D with RA can outperform the original variant, even considering different population sizes.

Moreover, this work aims to understand better the mechanics of MOEA/D regarding the population update. This behaviour suggests using RA techniques to mitigate common problems related to the choice of population size. Overall, our results indicate that MOEA/D with RA shows a hybrid behaviour in our experiments: MOEA/D with RA converges to the Pareto front at about the same speed as MOEA/D with a small population while exploring the search space as MOEA/D with a large population does. These results motivate the use of RA to mitigate common problems related to population size choice, independently of the MOP in question. Together these results provide important insights into the new population dynamics introduced by MOEA/D with RA that accelerates the convergence speed with additional improvements in the number of non-dominated solutions on the final approximation of the Pareto Front.

The remainder of this paper is organised as follows: Section \ref{section:RA} reviews the main concepts related to Resource Allocation in the MOEA/D. Section \ref{section:popsize} explains the relationship between population size and Resource Allocation in MOEA/D. Section \ref{chap3:moead_ps_explanation} describes the MOEA/D with partial update strategy (PS). Section \ref{section:experimental_setting} describes the experimental settings related to the investigation of the effect of partial updates on the performance of the algorithm, while Section \ref{section:results} presents the results of such investigation and comparisons between MOEA/D-PS against two different configurations of MOEA/D, one with small population size and another with big population size. Finally, Section \ref{section:conclusion} presents our concluding remarks.

\section{Resource Allocation}
\label{section:RA}

The key idea behind MOEA/D is to decompose a MOP into a set of single-objective sub-problems, which are then solved simultaneously. While these sub-problems are usually considered equivalent, a growing body of work indicates that prioritising some sub-problems at specific points of the search can improve the performance of MOEA/D. Techniques that address this issue are collectively called Resource Allocation (RA) techniques. 

Priority functions are used in RA to determine preferences between sub-problems. These functions take information about the progress of the search and return priority values that are then used to change the distribution of computational resources among sub-problems (\cite{cai2015external}). It is also possible to design priority functions that target other properties of the optimisation process (\cite{lavinas2019improving}), such as diversity or robustness (\cite{goulart2017robust}).

% Priority functions mediate the distribution of computational resources using a thresholding operation. At any given iteration $t$, let $u_i^t$ indicate the priority function value attributed to the $i$-th sub-problem, and $\upsilon^t$ be a threshold value. The subset of solutions selected for a variation on that iteration is defined as the sub-problems for which $u_i^t\geq \upsilon^t$.

% The original work on Resource Allocation for the MOEA/D \cite{zhang2009performance} defined a priority function known as the Relative Improvement (RI), defined as:

% \begin{equation}\label{sec2:ri}
%   u_i^t = \dfrac{f\left(\mathbf{x}_i^{t-\Delta t}
%   \right)-f\left(\mathbf{x}_i^t\right)}{f\left(\mathbf{x}_i^{t-\Delta t}\right)},
% \end{equation}

% where $f\left(\mathbf{x}_i^{t}\right)$ represents the aggregation function value of the incumbent solution to the $i$-th sub-problem on iteration $t$. $\Delta t$ is a parameter that controls how many generations to wait for the relative improvement comparison (notice that this definition assumes a minimisation problem and an aggregation function that always yields strictly positive values). 

\subsection{State of the art}
Much of the research on Resource Allocation has used the Relative Improvement (RI) as a priority function, with some modifications on other aspects of the algorithm (\cite{zhang2009performance,nasir2011improved}). Although Zhou et al. did expand the discussion over Resource Allocation in their work (\cite{zhou2016all}), few other works have studied Resource Allocation in depth. 

In previous works, we isolated the priority function as a point of investigation (\cite{lavinas2019improving,lavinas2019using}). The goal in these works was to improve the performance of MOEA/D based on the choice of priority function and to understand further the behaviour of MOEA/D under different Resource Allocation approaches. We introduced three new priority functions (Decision-Space Diversity, inverted Decision-Space Diversity and one based on random values). The experimental comparisons of these works revealed this surprising result: the performance of MOEA/D when using a random Resource Allocation method is as competitive as RI and better than not using Resource Allocation at all. This result suggested that MOEA/D may benefit simply from the increased populational inertia (possibly due to slower diversity loss) resulting from holding portions of the population constant during any given iteration.

Later, we investigated this question, using the Partial Update Strategy for the MOEA/D based on priority values sampled from a uniform distribution, leading to a random selection of sub-problems to update at every iteration (\cite{lavinas2020moea}). The Partial Update Strategy allows the control of the number of sub-problems modified at any given iteration and regulates the working population dynamics of the MOEA/D. The results reveal two main findings, and the first one is a strong association between more conservative updating of the MOEA/D population and improved performance. The second main finding is that MOEA/D benefits more from having slower population dynamics than from a specific prioritisation of sub-problems. These findings are in agreement with the results observed in the combinatorial domain by \cite{pruvost2020subproblemselection}. Therefore, we understand it is imperative to investigate the correspondences of MOEA/D with the Partial Update Strategy and the populational dynamics of MOEA/D with large and small population sizes.

\section{The Importance of Population Size}\label{section:popsize}

The population size is one of the essential parameters of Evolutionary Algorithms since it influences those algorithms' dynamics during the execution (\cite{glasmachers2014start}). While the right choice can lead to considerable improvements in the algorithm's performance, a wrong choice of population size might have the contrary effect (\cite{glasmachers2014start,pruvost2020subproblemselection}). For example, choosing a population size that is too small can cause premature convergence because a small population size might prevent the localization of optimal solutions. However, a bigger population size may cause the algorithm to waste computational resources, a critical issue in computationally costly problems (\cite{kohira2018proposal}). Furthermore, the proper choice of population size depends on the characteristics of different problems, such as the difficulty of the problem, the presence of multiple local-optima and the shape of the Pareto Front (\cite{vcrepinvsek2013exploration,glasmachers2014start,pruvost2020subproblemselection}). 

In MOEA/D, the population size also defines the number of sub-problems and the regions of the decision space where the sub-problems are established. Therefore, the choice of population size is even more critical for this algorithm. Since different decision space regions might have different characteristics, each sub-problem might display different characteristics in the same MOP. For example, given a MOP and two different sub-problems, while multiple local-optima might influence one sub-problem, another sub-problem might not have to deal with such difficulty. %This difference in the different sub-problems behaviour shows that choosing an adequate population size in MOEA/D is critical.

Considering all of this evidence, it is clear that choosing the right population size, especially in MOEA/D, is a delicate matter. That said, we reason that Resource Allocation can mitigate the burden of choosing the right population size in MOEA/D. That is because MOEA/D with Resource Allocation techniques benefits from maintaining a big population size and updating very few solutions at each iteration (\cite{pruvost2020subproblemselection, lavinas2020moea}), motivating the use of RA in MOEA/D. In other words, MOEA/D can maintain and improve a substantially large population at a lower cost than the original MOEA/D, the cost of updating those few selected solutions because RA techniques allow MOEA/D to update only a small subset of solutions from the whole larger population. By contrast, without RA techniques, a large population aggravates the computational burden (\cite{li2014evolutionary}), slows down the algorithm or introduces many non-useful sub-problems (\cite{nsga3_pt2}). Consequently, in this work, we aim to verify and understand the mechanics of MOEA/D regarding the population update with Resource Allocation techniques.

\section{The Partial Update Strategy}\label{chap3:moead_ps_explanation}

To verify and understand the mechanics of MOEA/D regarding the population update and investigate the extent of this effect, we use the Partial Update Strategy (PS), one of the most general Resource Allocation techniques. The PS selects $n$ individuals randomly to update each iteration, which is equivalent to a priority function sampled from a uniform distribution. In addition to that, the PS strategy always includes the boundary weight vectors' solutions, one for each objective. For example, there are two boundary weight vectors in a two objective MOP; thus, the number of sub-problems selected is $n + 2$. 

The reason for always selecting the boundaries is because they have an impact on the coordinates of the reference point $z^{*}$ used by the scalarizing function (\cite{wangNewResourceAllocation2019}). Algorithm~\ref{algo:moead_ps} details the pseudocode of the MOEA/D-DE (\cite{zhang2009performance}), using the Partial Update Strategy (MOEA/D-PS). Moreover, in a preliminary experiment, we saw a small (non-statistically significant) difference in performance favouring the case where the boundaries are always selected over the case where all vectors are selected randomly. Finally, we did not consider the case where the same set of the weight vectors is always used in all iterations, because~\cite{zhou2016all} showed an in-depth experiment that indicates that pre-specified vectors lead to worse performance to vectors specified on the run. 

\begin{algorithm}[htbp]
	\caption{MOEA/D-PS (MOEA/D-DE with the Partial Update Strategy)}\label{algo:moead_ps}
	\begin{algorithmic}[1]
		\State \textbf{Input}: number of subproblems to select \textbf{$n$}, termination criteria, number of objectives \textbf{$m$}. 
		\State \textbf{Initialize} MOEA/D-DE variables (e.g. weight vectors, set of solutions, etc.)
		\While{\textit{Termination criteria are not meet}}
		    \For {i = 1 to the number of subproblems}
                \State \textbf{${u_i} \leftarrow \text{rand()}$}\label{util_line} \Comment{Vector of random values}
            \EndFor
            \State\textbf{Select} $n+m$ subproblems: $n$ with the highest ${u_i}$ value and $m$ number of boundary subproblems.\label{select_line}
            % \State\textbf{Select} $n$ subproblems with the highest ${u_i}$ value.\label{select_line}
            % \State\textbf{Update} neighborhoods with selected subproblems.\label{update_line}
            \For {j = 1 to $n$} \Comment{Number of subproblems selected}\label{for_line}
		      %  \If{\textbf{\violet{subproblem $i$ was sampled}}}
	                \State \textbf{Generate} new candidate $y$ for subproblem $j$.\label{generate_line}
                    % \State \textbf{Generate} $n$ new candidates for selected subproblems.
                    \State  \textbf{Evaluate} these new candidate solutions.\label{evaluate_line}
                    % \State  \textbf{Evaluate} these $n$ new candidate solutions.
            %   \EndIf
		    \EndFor\label{endfor_line}
		  %\State\textbf{Update} neighborhoods with all subproblems.\label{update2_line}
		    \State \textbf{Update} the whole set of solutions with the $n$ new solutions generated in steps \ref{for_line}-\ref{endfor_line}.
		\EndWhile
	\end{algorithmic}
\end{algorithm}

Notice that the standard MOEA/D and other variants such as MOEA/D-DE can be instantiated from Algorithm \ref{algo:moead_ps} by setting $n = $ \textit{total number of sub-problems}. We highlight that this is only possible because we are using the generational version of MOEA/D. The only difference that the Partial Update Strategy introduces in the base algorithm is that only a few sub-problems are updated at any given iteration, regulated by the value of $n$. 

Other Resource Allocation techniques could also be expressed with this structure by modifying the priority value attribution function in Line \ref{util_line} of the algorithm. It is relevant to observe that sub-problems that are not selected by the Partial Update Strategy at a given iteration may still have their incumbent solutions updated. In other words, updating only small parts of the population using Resource Allocation in MOEA/D-PS affects only the variation step, not the replacement one. Thus, sub-problems not selected for variation may receive new candidate solutions, e.g., generated for a neighbouring sub-problem.

\subsection{Computational Complexity of MOEA/D-PS}

Here, we describe the additional computational complexity for allocating the computing resources in one iteration. On Line \ref{util_line} of Algorithm \ref{algo:moead_ps}, the priority value attribution has a complexity of $O(N)$, where $N$ is the population size. Similarly, the complexity for selecting sub-problems, on Line \ref{select_line}, is also $O(N)$. Thus, the additional complexity of one iteration caused by the Partial Update Strategy is $O(N)$, which is much less than the complexity of MOEA/D itself. While the reduction in the number of updated solutions increases the total number of generations, this reduction also decreases the number of evaluations per iteration. Thus, the total number of evaluations is kept exactly the same. 

According to this analysis, we can conclude that the additional computational complexity of MOEA/D-PS is minimal and that any increment in the performance of MOEA/D when using the Partial Update Strategy comes at a low cost.

% Here, we describe the additional computational complexity for allocating the computing resources. In Line \ref{util_line}, the priority value attribution has a complexity of $O(n)$. Similarly, the complexity for selecting sub-problems, in Line \ref{select_line}, is also $O(n)$. The major computational complexity introduced by the Partial Update Strategy is on Lines \ref{update_line} and \ref{update2_line}. The complexity for calculating the neighborhood relations on Line \ref{update_line} is $O(n(n - 1)/2$, distance calculations per iteration~\cite{moeadr_package, moeadr_paper}), given $n$ the number of sub-problems selected. The same calculation on Line \ref{update2_line} costs $O(N(N - 1 )/2)$, with $N$ being the population size. The cause for this increment in complexity is that the algorithm needs to calculate the neighbourhood relations at each iteration, since the selected sub-problems change across iterations (Line \ref{update_line}) for the iteration and evaluation of candidate solutions and then re-calculate these relations for the whole population (Line \ref{update2_line}), for the update process. Considering all the above considerations and computations, the additional complexity of one iteration caused by MOEA/D-PS is $O(N(N - 1 )/2)$, since $N > n$.

\section{Population Dynamics Behaviour Experiment}
\label{section:experimental_setting}

% We carry out the following experimental study to demonstrate the relationship between the partial update strategy (based on Resource Allocation) and population size. We compare the MOEA/D-PS against two configurations of MOEA/D with no Resource Allocation: (1) the first has a big population, the same size as the population size of MOEA/D-PS, and (2) the second has a small population, the same size as the number of solutions updated by MOEA/D-PS at each iteration. This experiment aims to show the hybrid behaviour of MOEA/D-PS, with shared characteristics of both MOEA/D configurations studied here.

We carry out the following experimental study to investigate the relationship between Resource Allocation, represented by MOEA/D-PS, and population size in practice. We compare the final approximation sets' quality based on (a) the final approximation hypervolume and proportion non-dominated of the final approximation sets, (b) the hypervolume anytime performance, (c) the empirical attainment function and (d) the statistical analysis based on the Wilcoxon Rank Sum Tests. This experiment aims to show the hybrid behaviour of MOEA/D-PS, with shared characteristics of both MOEA/D configurations studied here.

\subsection{Benchmark Problems}
Four different benchmark sets were used: (1) the scalable DTLZ set (\cite{deb2005scalable}), with 2 objectives, (2) the inverted scalable DTLZ benchmark (DTLZ$^{-1}$) set (\cite{ishibuchi2016performance}, with 2 objectives, (3) the UF set (\cite{zhang2008multiobjective}), with 2 and 3 objectives, and finally, (4) )the Black-Box Optimization Bi-Objective Benchmark test functions (\cite{tusar2016coco}). For all functions in all benchmark sets, we set the number of dimensions to $D = 40$\footnote{This is the maximum value one can use in the Black-Box Optimization Bi-Objective Benchmark test. For consistency, we keep this value fixed in all problems.}. The implementation of the test problems available from the \textit{smoof} package (\cite{smoof}) was used in all experiments\footnote{For the inverted DTLZ benchmark (DTLZ$^{-1}$) we modify the DTLZ implementation from the \textit{smoof} package in order to invert the problem set.}.

\begin{enumerate}
    \item The four DTLZ Benchmark functions we use have the following features (\cite{huband2006review}):

\begin{itemize}
    \item DTLZ1: Linear Pareto Front - unimodal;
    \item DTLZ2: Concave Pareto Front - unimodal;
    \item DTLZ3: Concave Pareto Front - multimodal;
    \item DTLZ4: Concave Pareto Front - unimodal.
    % \item DTLZ5: Degenerate Pareto Front - unimodal;
    % \item DTLZ6: Degenerate Pareto Front - unimodal;
    % \item DTLZ7: Disconnected Pareto Front with concave and convex portions - multimodal.
\end{itemize}
 
 \item The features of the four inverted DTLZ benchmark (DTLZ$^{-1}$) set (\cite{ishibuchi2016performance}), are:

\begin{itemize}
    \item DTLZ1$^{-1}$: Linear Pareto Front;
    \item DTLZ2$^{-1}$: Convex Pareto Front;
    \item DTLZ3$^{-1}$: Convex Pareto Front;
    \item DTLZ4$^{-1}$: Convex Pareto Front.
\end{itemize}

\item The UF Benchmark set is composed of ten unconstrained test problems with Pareto sets designed to be challenging to existing algorithms (\cite{li2019comparison}). Problems UF1-UF7 are two-objective MOPs, while UF8-UF10 are three-objective problems (\cite{zhang2008multiobjective}).

\begin{itemize}
    \item UF1: Convex Pareto Front - multimodal;
    \item UF2: Convex Pareto Front - multimodal;
    \item UF3: Convex Pareto Front - multimodal;
    \item UF4: Concave Pareto Front - multimodal;
    \item UF5: Linear Pareto Front - multimodal;
    \item UF6: Linear Pareto Front - multimodal;
    \item UF7: Linear Pareto Front - multimodal;
    \item UF8: Concave Pareto Front - multimodal;
    \item UF9: Linear and discontinuous Pareto Front - multimodal;
    \item UF10: Concave Pareto Front - multimodal.
\end{itemize}

\item Finally, we list the problems features of the Black-Box Optimization Bi-Objective Benchmark and then we describe the meaning of the features of these problems. We select all 55 two-objectives problems from this test suit.%, from the following groups:

\begin{itemize}
	\item F1, F2, F11: separable - separable;
    \item F3, F4, F12, F13: separable - moderate;
	\item F5, F6, F14, F15: separable - ill-conditioned;
	\item F7, F8, F16, F17: separable - multi-modal;
	\item F9, F10, F18, F19: separable - weakly-structured;
	\item F20, F21, F28: moderate - moderate;
	\item F22, F23, F29, F30: moderate - ill-conditioned;
	\item F24, F25, F31, F32: moderate - multi-modal;
	\item F26, F27, F33, F34: moderate - weakly-structured;
	\item F35, F36, F41: ill-conditioned - ill-conditioned;
	\item F37, F38, F42, F43: ill-conditioned - multi-modal;
	\item F39, F40, F44, F45: ill-conditioned - weakly-structured;
	\item F46, F47, F50: multi-modal - multi-modal;
	\item F48, F49, F51, F52: multi-modal - weakly structured;
	\item F53, F54, F55: weakly-structured - weakly-structured.
\end{itemize}
\end{enumerate}

Here, we give a simple explanation of the features in the problems above. The first feature is separability. A separable function does not show any dependencies between the variables. The next features are the moderate-conditioned and ill-conditioned, and they indicate how much changing a solution impacts the objective value (\cite{hansen2011impacts}). If the impact is high, we have an ill-conditioned function; otherwise, we have a moderate-conditioned function. A multi-modal function has at least two local optima, and finally, a weakly-structured function is a function that with a very unclear general structure (\cite{finck2010real}).

\subsection{Experimental Parameters}~\label{parameters}

We used the MOEA/D-DE parameters as they were introduced in the work of Li and Zhang (\cite{li2009multiobjective}) in all tests. All objectives were linearly scaled at every iteration to the interval $\left[0,1\right]$, and the Sobol decomposition (\cite{zapotecas-martinezLowdiscrepancySequencesTheir2015}) function was used.  Table (\ref{chap4:parameter_table}) summarizes the experimental parameters. Details of these parameters can be found in the documentation of package MOEADr and the original MOEA/D-DE reference (\cite{zhang2009performance,moeadr_package,moeadr_paper}). 

We highlight the choice of the $n$ parameter value of the Partial Update Strategy. This choice is based on our previous work (\cite{lavinas2020moea}), where the influence of different $n$ values on the performance of the MOEA/D-PS is investigated. The results demonstrated that low values are associated with (anytime) performance improvements, especially when $n$ is close to $10\%$ of the population size. Therefore, in this work, we set $n$ equal to $10\%$ of the population size, meaning a value of $n = 50$, since the population size is set to $500$\footnote{The number of selected sub-problems is equal to 50 + $m$, $m$ being the number of objectives of the MOP being solved.}.

Recently, it was pointed out that during the search, multi-objective algorithms might discard non-dominated solutions in favour of dominated ones (\cite{Multiobjective_Archiving}). In consequence, the final population of EMO algorithms might maintain a subset of solutions that are dominated by other solutions already discarded. In another work (\cite{ishibuchi_pang_shang_2020}), it was shown that a better final solution set, in comparison with the final population, can be obtained if this final solution set is constructed considering solutions from many, if not all, iterations. Therefore, we understand that such good solutions found by MOEA/D during the run might be excluded from the final population, a factor that is even more noticeable when MOEA/D maintains a small population size. To alleviate such problem, we use a bounded external archive with the same size as the population in use.

\begin{table}[htbp]
\centering
	\small
	\caption{Experimental parameter settings.}
	\label{chap4:parameter_table}
    \begin{tabular}{l|l}
        
        \rowcolor[gray]{.85}Parameters             & Value            \\ \hline
        DE mutation parameter                                  & $F = 0.25$       \\ \hline
        \multirow{2}{*}{Polynomial mutation parameters}        & $\eta_m = 20$    \\
                                                            & $p_m = 0.01$     \\ \hline
        Restricted Update parameter                            & $nr = 2$         \\ \hline
        Locality parameter                                  & $\delta_p = 0.9$ \\ \hline
        Neighborhood size                                   & $T = 20\%$ of the pop. size         \\ \hline
        Bounded External Archive & TRUE \\ 
        
        \multicolumn{2}{c}{}\\
        
        \rowcolor[gray]{.85} Decomposition method & Value \\ \hline
        Sobol \cite{zapotecas-martinezLowdiscrepancySequencesTheir2015} & 50 or 500 weights  \\ 
        
        \multicolumn{2}{c}{}\\
        
        \rowcolor[gray]{.85} Population size  & Value            \\ \hline
        MOEA/D-PS                                                  & 500     \\ \hline
        MOEA/D with big pop.                                       & 500    \\ \hline
        MOEA/D with small pop.                                     & 50     \\ 
        
        \multicolumn{2}{c}{}\\

        \rowcolor[gray]{.85}Resource allocation parameter  & Value            \\ \hline
        \multirow{2}{*}{$n$} & 50 + the boundary \\ 
        & weight vectors \\
        % 10\% of the pop. size \\
        % &  the boundary weight\\
        % & vectors    \\ \hline

        \multicolumn{2}{c}{}\\ 
        
        \rowcolor[gray]{.85}Experiment Parameters           & Value            \\ \hline
        Repeated runs                                       & 10               \\ \hline
        Computational budget                                & 100000 evals.            \\ 
\end{tabular}
\end{table}

\subsection{Experimental Evaluation}~\label{evaluation}

As indicated previously, we use the following criteria to compare the results of the different strategies: (a) final approximation hypervolume (HV) and proportion non-dominated solutions, (b) anytime performance, and (c) empirical attainment performance (EAP). For the calculation of the hypervolume, the objective function approximation values were scaled to the (0, 1) interval, with the reference point set to the nadir point: (1, 1) for two objective problems and (1, 1, 1) for three objective problems. We scale the approximation values using the range of objective values found by all algorithms in the comparisons. There is some concern that using a reference point equal to the nadir point may cause problems when calculating the hypervolume. However, because all problems have two or three objectives and the population size is large, this is not a serious issue for this experiment.

The hypervolume can be defined as the volume of the n-dimensional polygon formed by some reference point, $\vec{x}_{\text{ref}} = (x_1, x_2, ..., x_m)$ ,  and a finite set $S = (x_1, x_2, ..., x_m)$ of solutions in the positive orthant, $\mathbb{R}^{n}_{\geq 0}$, with $m$ being the number of objectives (\cite{beume2009complexity}). That is:

\begin{equation}
\centering
    HV(S) = \text{Hypercube calculate between} S \text{ and }\vec{x}_{\text{ref}}.
\end{equation}

\newpage

\paragraph{Final Approximation: Hypervolume and Proportion Non-dominated Solutions}~\label{archive_size}

We compare the final approximation hypervolume results of the algorithms in terms of hypervolume and the proportion non-dominated solutions (for both, higher is better). The proportion non-dominated solutions provides insights into how diverse is a set of solutions. The higher number of non-dominated solutions in a given set suggests a more diverse set of solutions in the objective space.

For a fair comparison, in special for the hypervolume results, the algorithms are evaluated based on an archive of $size = 500$. For MOEA/D-PS and MOEA/D with the big population, the archive is just the final population, but for MOEA/D with the small population, the archive accumulates solutions of different iterations. In specific, given that the small population is of size $50$, the archive of this variant combines solutions from $10$ iterations, lead to an archive of $size = 50 * 10 = 500$.

\paragraph{Anytime Hypervolume Performance}

Analysing MOP solvers using the final approximation provides limited information related to these algorithms' performance since any MOP should return a suitable set of solutions at any time during the search. Thus, it is imperative to extend the performance analysis because good solutions should be found as earlier as possible independently of the termination criterion or if the execution is interrupted during the search (\cite{zilberstein1996using,radulescu2013automatically,anytimePLS,tanabe2017benchmarking}). Here, we analyse the anytime performance effects in terms of HV values to investigate the impact of different population sizes in MOEA/D against MOEA/D-PS. As for the final approximation analysis, we use an archive of size $500$ for all algorithms.

% \paragraph{Inverted Generational Distance}
% Ishibuchi et al. ~\cite{ishibuchi2015modified} defines the Inverted Generational Distance (IGD) as the average distance from each reference point to its nearest solution.

% Mathematically, they define the IGD as follows:

% \begin{equation}
% \centering
%     IGD(S) = \frac{1}{|Z|} \sum\limits_{i=1}^{|Z|} \hat{d}_i,
% \end{equation}

% where $S = (x_1, x_2, ..., x_m)$, $m$ is the number of objectives, is a finite reference set of solutions in the positive orthant, $\mathbb{R}^{n}_{\geq 0}$, $Z = \{z_1, z_2,..., z_{|z|}\} $, is the reference set of solutions of size $n$, and $\hat{d}$ is the Euclidean distance from $z_j$, $j \in 1,..., n$ to its nearest objective solution in $S$.

% When the IGD indicator returns a small value, it suggests that the set of solutions evaluated have both good convergence and diversity, provided with a well-distributed reference set of solutions~\cite{ishibuchi2015modified}.

\paragraph{Empirical Attainment Performance}

The empirical attainment function (EAF) allows the examination of the solution many sets of different runs of an algorithm, and it can illustrate where and by how much the outcomes of two algorithms differ in the objective space (\cite{lopez2010exploratory}). The EAF is based on the attainment surface \textit{and represents the probability that an arbitrary objective region in the objective space is attained (that is, dominated or equal) by an algorithm and probability can be estimated using data collected from several independent runs of such algorithm.} The attained surface separates the objective space into two regions: (1) where the objective space is dominated (attained) by solutions of many sets and (2) where the objective space is not dominated by those solutions (\cite{fonseca1996performance,da2001inferential}). For example, the median attainment surface shows regions dominated by at least half of the runs (\cite{lopez2010exploratory}). 

\paragraph{Statistical Analysis}

We also evaluate the HV performance statistically in terms of the final (archive) population. The techniques' differences are analysed using Wilcoxon Rank Sum Tests, with a  significance level of $\alpha = 0.05$ and Hommel adjustment for multiple comparisons. 

\subsection{Reproducibility}

For reproducibility purposes, all the code and experimental scripts are available online at \href{https://github.com/yurilavinas/MOEADr/tree/ECJ}{https://github.com/yurilavinas/MOEADr/tree/ECJ}. Futhermore, the data for all experiments is available at \href{https://zenodo.org/record/5551197#.YV6EE6CRVhF}{https://zenodo.org/record/5551197\#.YV6EE6CRVhF}.
\section{Results}~\label{section:results}

The following Section of this paper moves on to describe in greater detail the experimental results in terms of (a) the final approximation hypervolume and the proportion non-dominated of the final approximation sets, (b) the hypervolume anytime performance, (c) the empirical attainment function and (d) the statistical analysis based on the Wilcoxon Rank Sum Tests. These analytical procedures and the results obtained from them are described in the previous Section~\ref{section:experimental_setting}.

\subsection{Final Approximation Results}

Table \ref{chap5:stats_bibbob_hv} shows, on the left side, the mean results obtained by the MOEA/D-PS with $n = 50$ and MOEA/D without Resource Allocation with (1) small population size and (2) big population size for all test problems. We recall that for fair of comparison, we use an archive population of $size = 500$ for all algorithms, see Section~\ref{archive_size}.

%Table \ref{chap5:pvals} presents the results of statistical pairwise comparisons using the same methodology described in subsection~\ref{evaluation}, corroborating the results observed in Table \ref{chap5:stats_bibbob_hv}.

% \subsection{Proportion of non-dominated Solutions of the Final Approximation}\label{chap4:nndom}

\begin{table}[tbp]
\vspace{-2.85em}
\centering
\scriptsize
\caption[Means and standard errors for HV.]{Means (standard errors) for HV and proportion of non-dominated solutions, for each problem. Best results are highlighted in bold.}
\vspace{-1.55em}
\label{chap5:stats_bibbob_hv}
\begin{tabular}[t]{c||c|c|c||c|c|c}
    
    \rowcolor[gray]{.8} \multicolumn{4}{c||}{\textbf{HV}} &\multicolumn{3}{c}{\textbf{NNDOM}}\\ \hline 
    \rowcolor[gray]{.9} MOP & MOEA/D-PS & Big population & Small population & MOEA/D-PS & Big population & Small population \\ \hline 
% \rowcolor[gray]{.95} DTLZ problems & MOEA/D-PS = 50 & Big pop. & Small pop. \\ \hline 
DTLZ1              & \textbf{1.00 (0.01)} &\textbf{1.00 (0.00)} & \textbf{1.00 (0.00)} & \textbf{1.00 (0.00) }&0.13 (0.08) &0.66 (0.25)\\
DTLZ2              & \textbf{0.99 (0.00)} &\textbf{0.99 (0.00)} &\textbf{0.99 (0.00)} &\textbf{0.99 (0.02)} &0.95 (0.02) &0.92 (0.05)\\
DTLZ3              & \textbf{1.00 (0.01)} &\textbf{1.00 (0.00)} &\textbf{1.00 (0.00)} &\textbf{0.99 (0.01)} &0.11 (0.13) &0.80 (0.20)\\
DTLZ4              & \textbf{0.99 (0.00)} &\textbf{0.99 (0.00)} &\textbf{0.99 (0.00)} &\textbf{0.99 (0.02)} &0.88 (0.05) &0.91 (0.05)\\

\rowcolor[gray]{.95}DTLZ1$^{-1}$ & 0.44 (0.02) &0.35 (0.01) &\textbf{0.46 (0.01)}& \textbf{1.00 (0.00)} &0.29 (0.04) &0.98 (0.02)\\
\rowcolor[gray]{.95}DTLZ2$^{-1}$          & \textbf{0.78 (0.00)} &\textbf{0.78 (0.00)} &0.77 (0.00)& \textbf{1.00 (0.00)} &\textbf{1.00 (0.00)} &\textbf{1.00 (0.01)}\\
\rowcolor[gray]{.95}DTLZ3$^{-1}$ & \textbf{0.75 (0.01)} &0.60 (0.03) & \textbf{0.75 (0.01)}& \textbf{1.00 (0.00)} &0.17 (0.03) &0.96 (0.03)\\
\rowcolor[gray]{.95}DTLZ4$^{-1}$          & \textbf{0.78 (0.00)} &\textbf{0.78 (0.00)} &0.77 (0.00)& \textbf{1.00 (0.00)} &\textbf{1.00 (0.00)} &0.99 (0.01)\\

UF1       & 0.94 (0.02) &\textbf{0.98 (0.00)} &0.95 (0.02) & \textbf{0.96 (0.04)} &0.37 (0.06) &0.54 (0.07)\\
UF2       & 0.92 (0.01) &\textbf{0.94 (0.01)} &\textbf{0.94 (0.01)} & \textbf{0.98 (0.03)} & 0.38 (0.05) &0.54 (0.04)\\
UF3       & 0.91 (0.03) &\textbf{0.98 (0.00)} &0.96 (0.03) & \textbf{0.97 (0.04)} &0.24 (0.03) &0.53 (0.09)\\
UF4       & 0.48 (0.00) &\textbf{0.50 (0.00)} &0.47 (0.01) & \textbf{0.96 (0.04)} &0.56 (0.05) &0.73 (0.07)\\
UF5       & 0.90 (0.04) &\textbf{0.96 (0.01)} &0.90 (0.03) & \textbf{0.78 (0.20)} &0.20 (0.03) &0.44 (0.10)\\
UF6       & 0.92 (0.04) &\textbf{0.97 (0.00)} &0.93 (0.05) & \textbf{0.96 (0.04)} &0.31 (0.05) &0.50 (0.06)\\
UF7       & 0.92 (0.08) &\textbf{0.97 (0.00)} &0.94 (0.06) & \textbf{0.93 (0.21)} &0.48 (0.08) &0.61 (0.11)\\
UF8                & 0.99 (0.00) &\textbf{1.00 (0.00)} &0.99 (0.00) & \textbf{0.99 (0.01)} &0.64 (0.09) &0.79 (0.05)\\
UF9                & \textbf{0.99 (0.00)} &\textbf{0.99 (0.00)} &\textbf{0.99 (0.00)} & \textbf{0.99 (0.01)} &0.60 (0.12) &0.70 (0.10)\\
UF10      & \textbf{0.98 (0.01)} &\textbf{0.98 (0.00)} &0.97 (0.01) & \textbf{0.93 (0.08)} &0.46 (0.04) &0.73 (0.08)\\

\rowcolor[gray]{.95}F1         & \textbf{0.96 (0.00)} &\textbf{0.96 (0.01)} &\textbf{0.96 (0.00)} & \textbf{0.99 (0.01)} &0.32 (0.06) &0.50 (0.04)\\
\rowcolor[gray]{.95}F2         & \textbf{1.00 (0.01)} &0.99 (0.00) &\textbf{1.00 (0.00)} & \textbf{0.99 (0.01)} &0.29 (0.06) &0.54 (0.08)\\
\rowcolor[gray]{.95}F3                  & \textbf{0.98 (0.00)} &\textbf{0.98 (0.00)} &\textbf{0.98 (0.00)} & \textbf{1.00 (0.01)} &0.36 (0.05) &0.58 (0.04)\\
\rowcolor[gray]{.95}F4                  & \textbf{1.00 (0.00)} &0.99 (0.00) &\textbf{1.00 (0.00)} & \textbf{0.99 (0.01)} &0.25 (0.05) &0.47 (0.07)\\
\rowcolor[gray]{.95}F5         & 0.91 (0.01) &0.90 (0.01) &\textbf{0.93 (0.00)} & \textbf{0.99 (0.01)} &0.33 (0.03) &0.49 (0.06)\\
\rowcolor[gray]{.95}F6                  & 0.99 (0.00) &0.99 (0.00) &\textbf{1.00 (0.00)} & \textbf{0.99 (0.01)} &0.25 (0.05) &0.51 (0.06)\\
\rowcolor[gray]{.95}F7                  & 0.99 (0.00) &0.99 (0.00) &\textbf{1.00 (0.00)} & \textbf{0.99 (0.01)} &0.18 (0.02) &0.37 (0.07)\\
\rowcolor[gray]{.95}F8                  & \textbf{0.99 (0.00)} &\textbf{0.99 (0.00)} &\textbf{0.99 (0.00)} & \textbf{0.89 (0.11)} &0.16 (0.01) &0.42 (0.11)\\
\rowcolor[gray]{.95}F9                  & \textbf{1.00 (0.00)} &0.99 (0.00) &\textbf{1.00 (0.00)} & \textbf{0.99 (0.01)} &0.31 (0.06) &0.54 (0.09)\\
\rowcolor[gray]{.95}F10        & 0.79 (0.03) &\textbf{0.80 (0.00)} &0.77 (0.05) & \textbf{0.99 (0.01)} &0.45 (0.07) &0.55 (0.05)\\
\rowcolor[gray]{.95}F11                 & \textbf{0.98 (0.00)} &\textbf{0.98 (0.00)} &\textbf{0.98 (0.00)} & \textbf{0.99 (0.01)} &0.31 (0.06) &0.55 (0.06)\\
\rowcolor[gray]{.95}F12                          & \textbf{1.00 (0.00)} &\textbf{1.00 (0.00)} &\textbf{1.00 (0.00)} & \textbf{0.96 (0.04)} &0.29 (0.05) &0.62 (0.06)\\
\rowcolor[gray]{.95}F13                          & \textbf{1.00 (0.00)} &\textbf{1.00 (0.00)} &\textbf{1.00 (0.00)} & \textbf{0.99 (0.01)} &0.27 (0.05) &0.53 (0.07)\\
\rowcolor[gray]{.95}F14        & 0.95 (0.01) &0.95 (0.01) &\textbf{0.98 (0.01)} & \textbf{0.99 (0.02)} &0.30 (0.05) &0.48 (0.05)\\
\rowcolor[gray]{.95}F15                          & \textbf{1.00 (0.00)} &\textbf{1.00 (0.00)} &\textbf{1.00 (0.00)} & \textbf{0.99 (0.01)} &0.28 (0.04) &0.49 (0.05)\\
\rowcolor[gray]{.95}F16                 & 0.99 (0.00) &\textbf{1.00 (0.00)} &0.99 (0.00) & \textbf{0.99 (0.01)} &0.17 (0.01) &0.41 (0.14)\\
\rowcolor[gray]{.95}F17        & 0.99 (0.01) &\textbf{1.00 (0.00)} &\textbf{1.00 (0.00)} & \textbf{0.94 (0.05)} &0.15 (0.02) &0.39 (0.10)\\
\rowcolor[gray]{.95}F18                          & \textbf{1.00 (0.00)} &\textbf{1.00 (0.00)} &\textbf{1.00 (0.00)} & \textbf{0.99 (0.01)} &0.31 (0.04) &0.54 (0.09)\\
\rowcolor[gray]{.95}F19        & 0.93 (0.07) &\textbf{0.96 (0.02)} &0.93 (0.04) & \textbf{0.99 (0.02)} &0.40 (0.07) &0.62 (0.08)\\
\rowcolor[gray]{.95}F20                          & \textbf{1.00 (0.00)} &\textbf{1.00 (0.00)} &\textbf{1.00 (0.00)} & \textbf{0.98 (0.02)} &0.43 (0.12) &0.66 (0.07)\\
\rowcolor[gray]{.95}F21                          & \textbf{1.00 (0.00)} &\textbf{1.00 (0.00)} &\textbf{1.00 (0.00)} & \textbf{0.99 (0.02)} &0.28 (0.06) &0.56 (0.07)\\
\rowcolor[gray]{.95}F22        & 0.92 (0.03) &\textbf{0.96 (0.01)} &\textbf{0.96 (0.02)} & \textbf{0.99 (0.02)} &0.33 (0.06) &0.55 (0.06)\\
\rowcolor[gray]{.95}F23                          & \textbf{1.00 (0.00)} &\textbf{1.00 (0.00)} &\textbf{1.00 (0.00)} & \textbf{0.99 (0.01)} &0.31 (0.04) &0.54 (0.04)\\
\rowcolor[gray]{.95}F24                 & 0.99 (0.00) &\textbf{1.00 (0.00)} &\textbf{1.00 (0.00)} & \textbf{0.96 (0.04)} &0.16 (0.02) &0.39 (0.07)\\
\rowcolor[gray]{.95}F25                 & 0.99 (0.00) &\textbf{1.00 (0.00)} &0.99 (0.00) & \textbf{0.92 (0.10)} &0.14 (0.02) &0.32 (0.07)\\
\rowcolor[gray]{.95}F26                          & \textbf{1.00 (0.00)} &\textbf{1.00 (0.00)} &\textbf{1.00 (0.00)} & \textbf{0.98 (0.02)} &0.37 (0.10) &0.70 (0.13)\\
\rowcolor[gray]{.95}F27        & 0.90 (0.07) &\textbf{0.94 (0.02)} &0.90 (0.06) & \textbf{0.99 (0.01)} &0.37 (0.05) &0.62 (0.06)\\
\rowcolor[gray]{.95}F28                          & \textbf{1.00 (0.00)} &\textbf{1.00 (0.00)} &\textbf{1.00 (0.00)} & \textbf{0.99 (0.02)} &0.24 (0.06) &0.47 (0.08)\\
\rowcolor[gray]{.95}F29        & 0.96 (0.01) &0.96 (0.01) &\textbf{0.98 (0.00)} & \textbf{0.99 (0.02)} &0.24 (0.05) &0.44 (0.05)\\
\rowcolor[gray]{.95}F30                          & \textbf{1.00 (0.00)} &\textbf{1.00 (0.00)} &\textbf{1.00 (0.00)} & \textbf{0.99 (0.02)} &0.27 (0.05) &0.45 (0.06)\\

\rowcolor[gray]{.95}F31        & 0.98 (0.01) &\textbf{1.00 (0.00) }&0.99 (0.00)  & \textbf{0.96 (0.07)} &0.16 (0.03) &0.42 (0.12)\\
\rowcolor[gray]{.95}F32                 & 0.99 (0.00) &\textbf{1.00 (0.00)} &0.99 (0.00)  & \textbf{0.95 (0.06)} &0.14 (0.02) &0.34 (0.06)\\
\rowcolor[gray]{.95}F33                          & \textbf{1.00 (0.00)} &\textbf{1.00 (0.00)} &\textbf{1.00 (0.00)} & \textbf{0.99 (0.02)} &0.25 (0.03) &0.56 (0.12)\\
\rowcolor[gray]{.95}F34        & 0.94 (0.02) &0.94 (0.01) &\textbf{0.95 (0.02)} & \textbf{0.99 (0.02)} &0.35 (0.11) &0.50 (0.05)\\
\rowcolor[gray]{.95}F35        & 0.78 (0.02) &0.81 (0.01) &\textbf{0.82 (0.01)} & \textbf{0.99 (0.01)} &0.40 (0.06) &0.55 (0.04)\\
\rowcolor[gray]{.95}F36        & 0.91 (0.03) &\textbf{0.97 (0.01)} &0.98 (0.02) & \textbf{0.99 (0.01)} &0.30 (0.04) &0.52 (0.05)\\
\rowcolor[gray]{.95}F37        & 0.89 (0.02) &0.96 (0.01) &\textbf{0.97 (0.01)} & \textbf{0.98 (0.02)} &0.21 (0.03) &0.42 (0.03)\\
\rowcolor[gray]{.95}F38        & 0.91 (0.03) &0.96 (0.02) &\textbf{0.97 (0.02)} & \textbf{0.91 (0.12)} &0.18 (0.05) &0.39 (0.06)\\
\rowcolor[gray]{.95}F39        & 0.94 (0.01) &0.95 (0.01) &\textbf{0.97 (0.01)} & \textbf{0.99 (0.01)} &0.33 (0.05) &0.53 (0.07)\\
\rowcolor[gray]{.95}F40        & 0.61 (0.05) &\textbf{0.67 (0.01)} &0.64 (0.04) & \textbf{0.99 (0.01)} &0.52 (0.09) &0.67 (0.05)\\

\rowcolor[gray]{.95}F41                          & \textbf{1.00 (0.00)} &\textbf{1.00 (0.00)} &\textbf{1.00 (0.00)} & \textbf{0.99 (0.01)} &0.26 (0.05) &0.50 (0.07)\\
\rowcolor[gray]{.95}F42        & 0.98 (0.01) &\textbf{1.00 (0.00)} &0.99 (0.00) & \textbf{0.98 (0.02)} &0.16 (0.03) &0.37 (0.05)\\
\rowcolor[gray]{.95}F43        & 0.99 (0.00) &\textbf{1.00 (0.00)} &\textbf{1.00 (0.01)} & \textbf{0.95 (0.06)} &0.15 (0.02) &0.35 (0.03)\\
\rowcolor[gray]{.95}F44                          & \textbf{1.00 (0.00)} &\textbf{1.00 (0.00) }&\textbf{1.00 (0.00)} & \textbf{0.99 (0.01)} &0.28 (0.05) &0.51 (0.10)\\
\rowcolor[gray]{.95}F45        & 0.93 (0.06) &\textbf{0.98 (0.02)} &0.95 (0.05) & \textbf{0.99 (0.01)} &0.36 (0.07) &0.58 (0.08)\\
\rowcolor[gray]{.95}F46        & 0.97 (0.01) &\textbf{1.00 (0.00)} &0.98 (0.01) & \textbf{0.99 (0.01)} &0.16 (0.03) &0.58 (0.15)\\
\rowcolor[gray]{.95}F47                 & 0.98 (0.00) &\textbf{1.00 (0.00)} &0.98 (0.00) & \textbf{0.99 (0.01)} &0.15 (0.03) &0.59 (0.15)\\
\rowcolor[gray]{.95}F48        & 0.98 (0.00) &\textbf{1.00 (0.00)} &0.98 (0.01) & \textbf{0.93 (0.13)} &0.15 (0.02) &0.39 (0.12)\\
\rowcolor[gray]{.95}F49        & 0.91 (0.05) &\textbf{0.97 (0.00)} &0.95 (0.03) & \textbf{0.98 (0.02)} &0.21 (0.03) &0.43 (0.10)\\
\rowcolor[gray]{.95}F50                 & 0.99 (0.00) &\textbf{1.00 (0.00)} &0.99 (0.00) & \textbf{0.99 (0.01)} &0.13 (0.02) &0.62 (0.10)\\
\rowcolor[gray]{.95}F51                 & \textbf{1.00 (0.00)} &\textbf{1.00 (0.00)} &\textbf{1.00 (0.01)} & \textbf{0.97 (0.03)} &0.14 (0.02) &0.36 (0.08)\\
\rowcolor[gray]{.95}F52        & 0.90 (0.05) &\textbf{0.98 (0.01)} &0.97 (0.03) & \textbf{0.98 (0.02)} &0.16 (0.03) &0.44 (0.09)\\
\rowcolor[gray]{.95}F53                          & \textbf{1.00 (0.00)} &\textbf{1.00 (0.00)} &\textbf{1.00 (0.00)} & \textbf{1.00 (0.01)} &0.37 (0.08) &0.61 (0.07)\\
\rowcolor[gray]{.95}F54        & 0.94 (0.06) &\textbf{0.97 (0.02)} &0.95 (0.03) & \textbf{1.00 (0.01)} &0.35 (0.05) &0.60 (0.06)\\
\rowcolor[gray]{.95}F55        & 0.22 (0.03) &\textbf{0.32 (0.04)} &0.22 (0.07) & \textbf{0.99 (0.01)} &0.87 (0.05) &0.95 (0.03)\\

\hline
Best/total &  31/73 & 57/73 & 43/73  &  73/73 & 2/73 & 1/73 \\
% \hline
\end{tabular}
\end{table}

Furthermore, looking at the proportion of non-dominated solutions in the right side of Table~\ref{chap5:stats_bibbob_hv}, we can see that updating a subset of solutions from a larger population at each iteration resulted in the highest proportion of non-dominated solutions on all MOPs. The main exception is on the UF benchmark set, where MOEA/D with a big population has a higher proportion non-dominated solutions (but not the highest proportion and duplicated non-dominated solutions). This higher overall number of non-dominated solutions indicates that the approximation set is much more diverse compared to the approximation sets of the traditional MOEA/D, independently of the population configuration. It is clear from both the hypervolume and the number of non-dominated solutions results that MOEA/D-PS that has a similar performance with a more diverse set of solutions in comparison with MOEA/D with big population size and better than MOEA/D with small population size.~\footnote{From results (available in the GitHub repository) obtained from our experiments we saw that MOEA/D-PS tends to group some solutions in the same region of the search space. Yet, in most MOPs studied here, this MOEA/D variant is still able to find more unique solutions than MOEA/D with a big population. We also noted that the number of unique solutions tends to decrease with the difficulty of the problem in question.}

\subsection{Anytime Performance}\label{chap4:anytime}

\begin{figure}[htbp]
    \centering
    \includegraphics[height=1\textwidth]{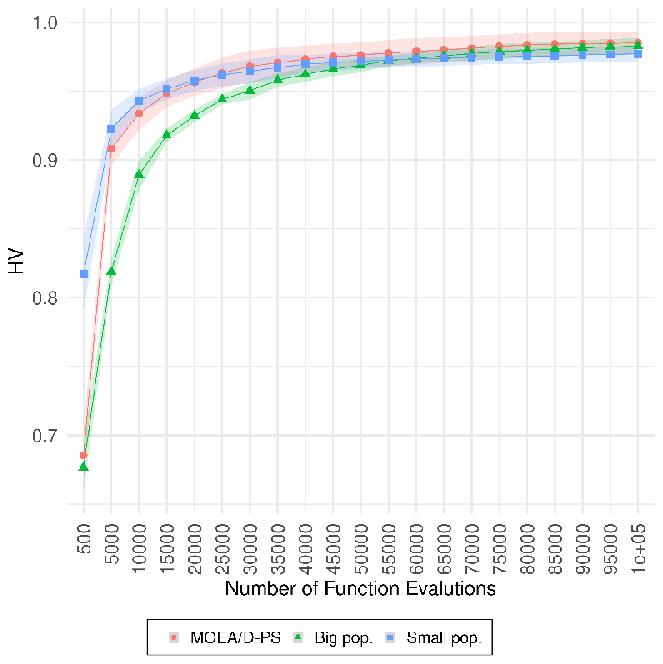}
    \caption{Anytime HV (higher is better, shaded areas indicate the standard deviation) on UF10. The performance of MOEA/D-PS is shown as the red circles, MOEA/D with population size 500 is shown as the green triangles, and MOEA/D with population size equals to 50 is shown as the blue squares, on UF10. The anytime performance of MOEA/D-PS is similar to the anytime performance of MOEA/D with a small population. We can see that the three variants have almost the same performance at the end of the search.}
    \label{fig:anytime_hv_uf10}
\end{figure}

\begin{figure}[htbp]
    \centering
    \includegraphics[width=1\textwidth]{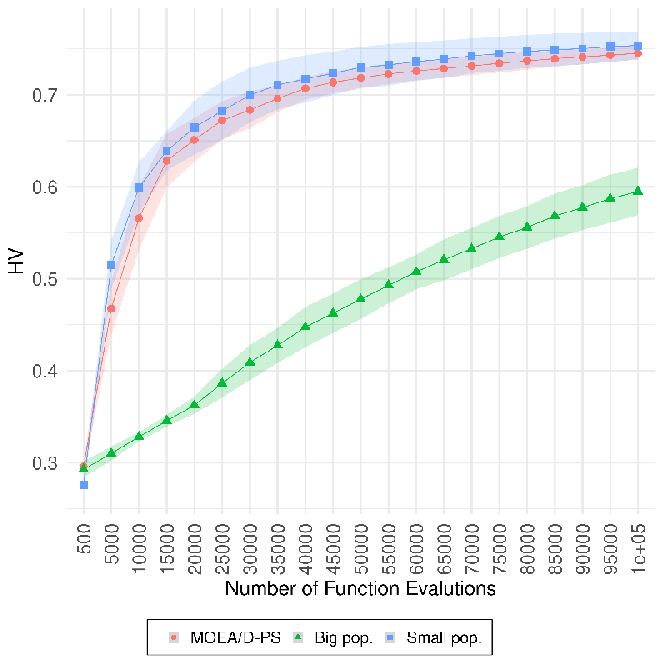}
    \caption{Anytime HV (higher is better, shaded areas indicate the standard deviation) on DTLZ3$^{-1}$. The performance of MOEA/D-PS is shown as the red circles, MOEA/D with population size 500 is shown as the green triangles, and MOEA/D with population size equals to 50 is shown as the blue squares, on DTLZ3$^{-1}$. The anytime performance of MOEA/D-PS is similar to the anytime performance of MOEA/D with a small population and much faster than MOEA/D with big population.}
    \label{fig:anytime_hv_inv_DTLZ3}
\end{figure}

\begin{figure}[htbp]
    \centering
    \includegraphics[width=1\textwidth]{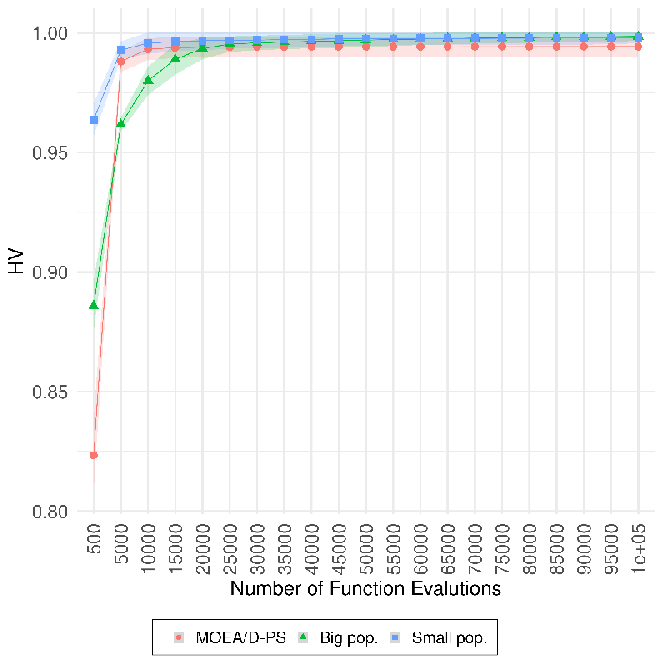}
    \caption{Anytime HV (higher is better, shaded areas indicate the standard deviation) on DTLZ1. The performance of MOEA/D-PS is shown as the red circles, MOEA/D with population size 500 is shown as the green triangles, and MOEA/D with population size equals to 50 is shown as the blue squares, on DTLZ1. The anytime performance of MOEA/D-PS is similar to the anytime performance of MOEA/D with a small population.}
    \label{fig:anytime_hv_DTLZ2}
\end{figure}

\begin{figure}[htbp]
    \centering
    \includegraphics[width=1\textwidth]{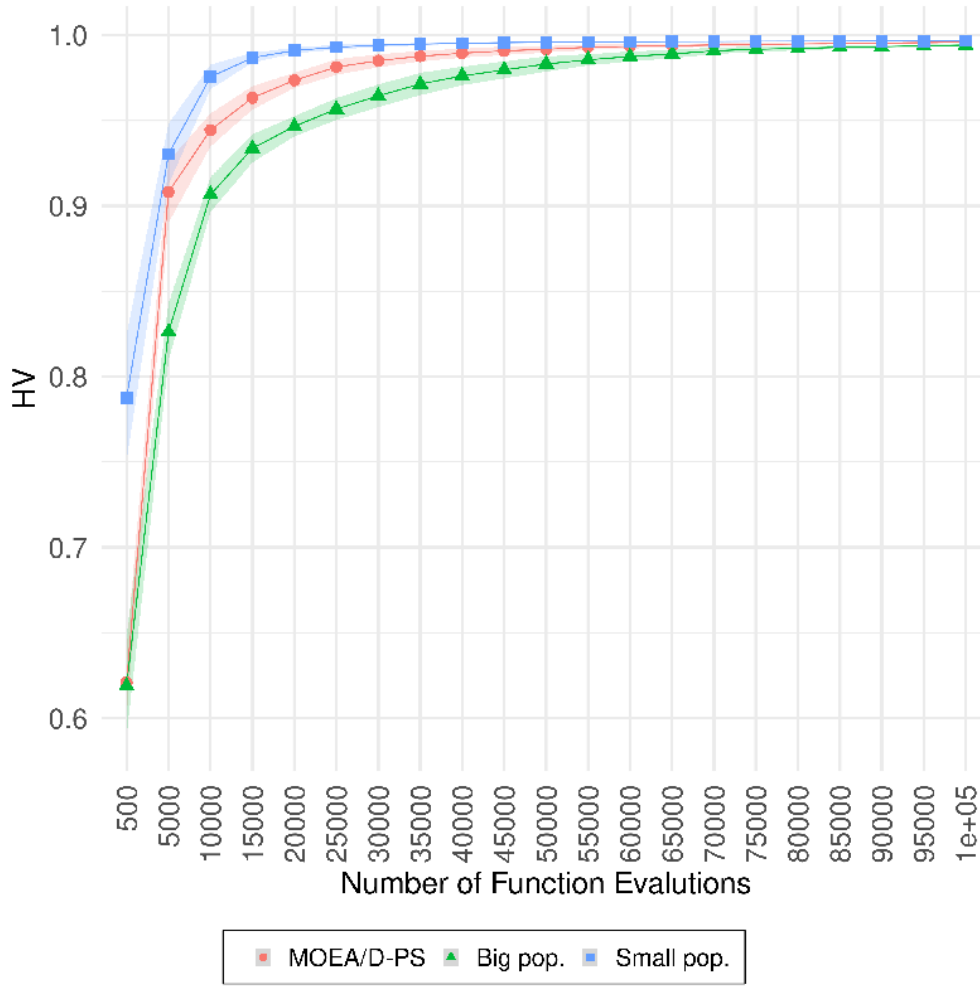}
    \caption{Anytime HV (higher is better, shaded areas indicate the standard deviation) on F9. The performance of MOEA/D-PS is shown as the red circles, MOEA/D with population size 500 is shown as the green triangles, and MOEA/D with population size equals to 50 is shown as the blue squares, on F9. The anytime performance of MOEA/D-PS shows that this algorithm converges as fast as MOEA/D with a small population and that MOEA/D is faster than MOEA/D with big population.}
    \label{fig:anytime_hv_F9}
\end{figure}

\begin{figure}[htbp]
\centering

\includegraphics[width=1\textwidth]{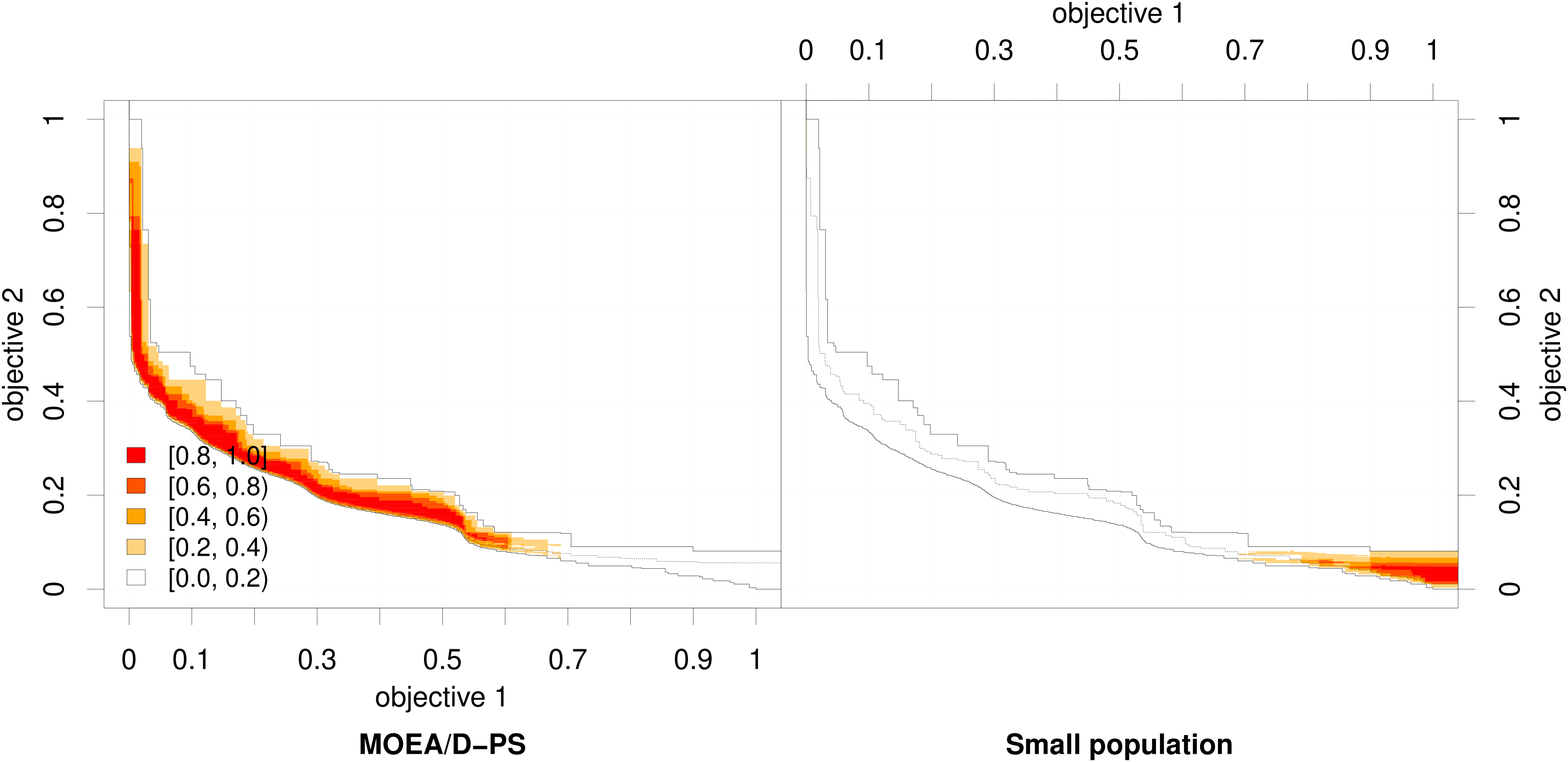}
\caption{Differences between the EAFs of MOEA/D-PS with MOEA/D with small population. The red shade encodes the magnitude of the observed difference on F22. The darker the red shade, the better performance of the algorithm. MOEA/D-PS (left) performs better in almost all regions of the objective space, except for the region closer to objective 1 (low values on the x-axis). This region is where MOEA/D with small population (right) performs better.}
\label{fig:eaf_F22_small}
\end{figure}
    % ~~
    % \qquad

\begin{figure}[htbp]
\centering
\includegraphics[width=1\textwidth]{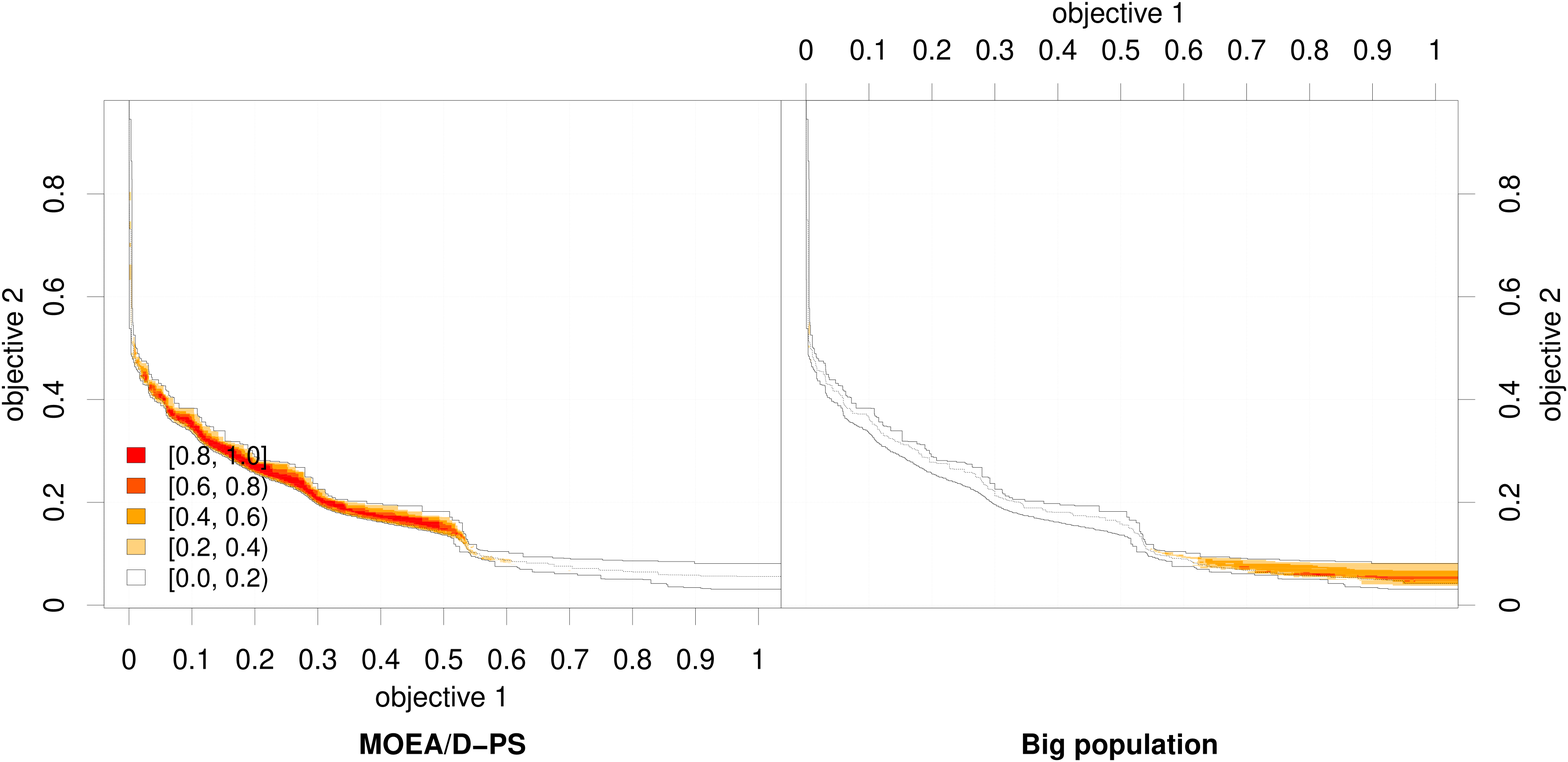}

	\caption{Differences between the EAFs of MOEA/D-PS with MOEA/D with big population. The red shade encodes the magnitude of the observed difference on F22. The darker the red shade, the better performance of the algorithm. MOEA/D-PS (left) performs better at the central regions of the objective space, with minimal advantage on the region closer to objective 2 (low values on the y-axis). MOEA/D with big population (right) performs better in the region closer to objective 1 (low values on the x-axis).}
	\label{fig:eaf_F22_big}
\end{figure}

\begin{figure}[htbp]
\centering
\includegraphics[width=1\textwidth]{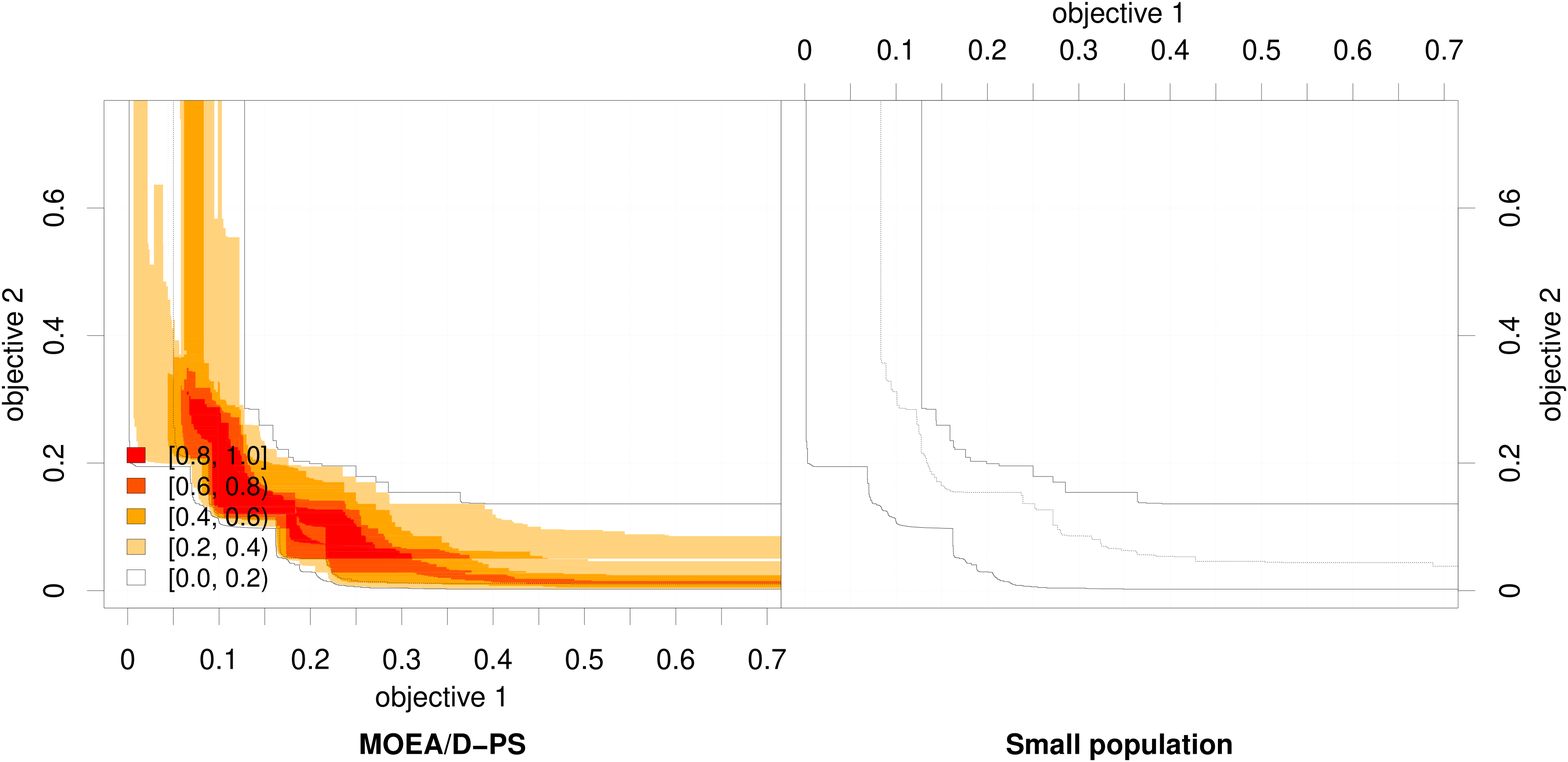}
\caption{Differences between the EAFs of MOEA/D-PS with MOEA/D with small population. The red shade encodes the magnitude of the observed difference on UF6. The darker the red shade, the better performance of the algorithm. MOEA/D-PS (left) performs well towards all regions of the objective space.}

\label{fig:eaf_UF6_small}
\end{figure}
    % ~~
    % \qquad
\begin{figure}[htbp]
\centering

\includegraphics[width=1\textwidth]{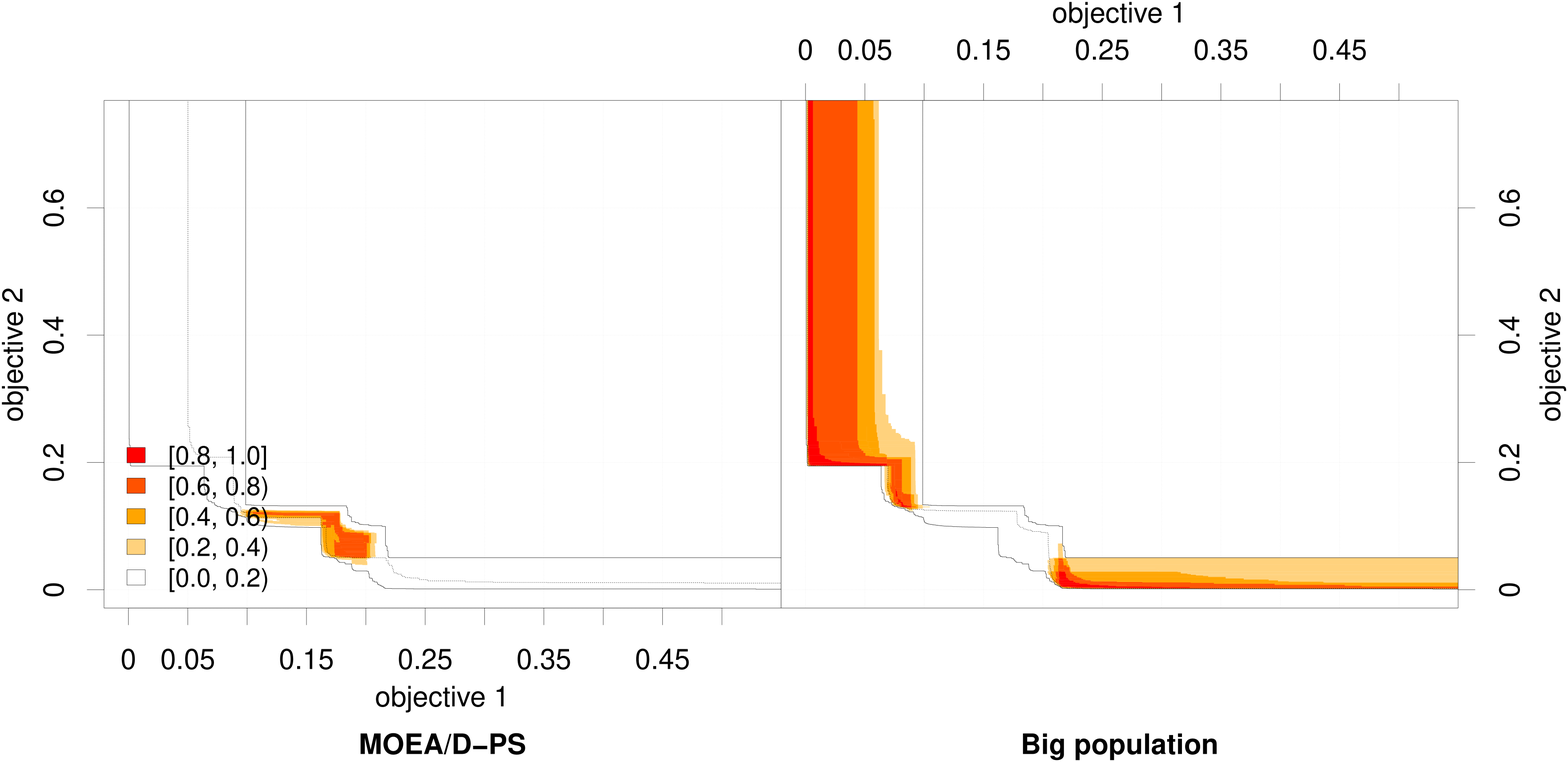}
\caption{Differences between the EAFs of MOEA/D-PS with MOEA/D with big population. The red shade encodes the magnitude of the observed difference on UF6. The darker the red shade, the better performance of the algorithm. MOEA/D-PS (left) performs better at central regions of the objective space. In contrast, MOEA/D with big population (right) performs better in the regions closer to objectives 1 and 2 (low values on the x-axis and y-axis.}

	\label{fig:eaf_UF6_big}
\end{figure}

\begin{figure}[htbp]
\centering
\includegraphics[width=1\textwidth]{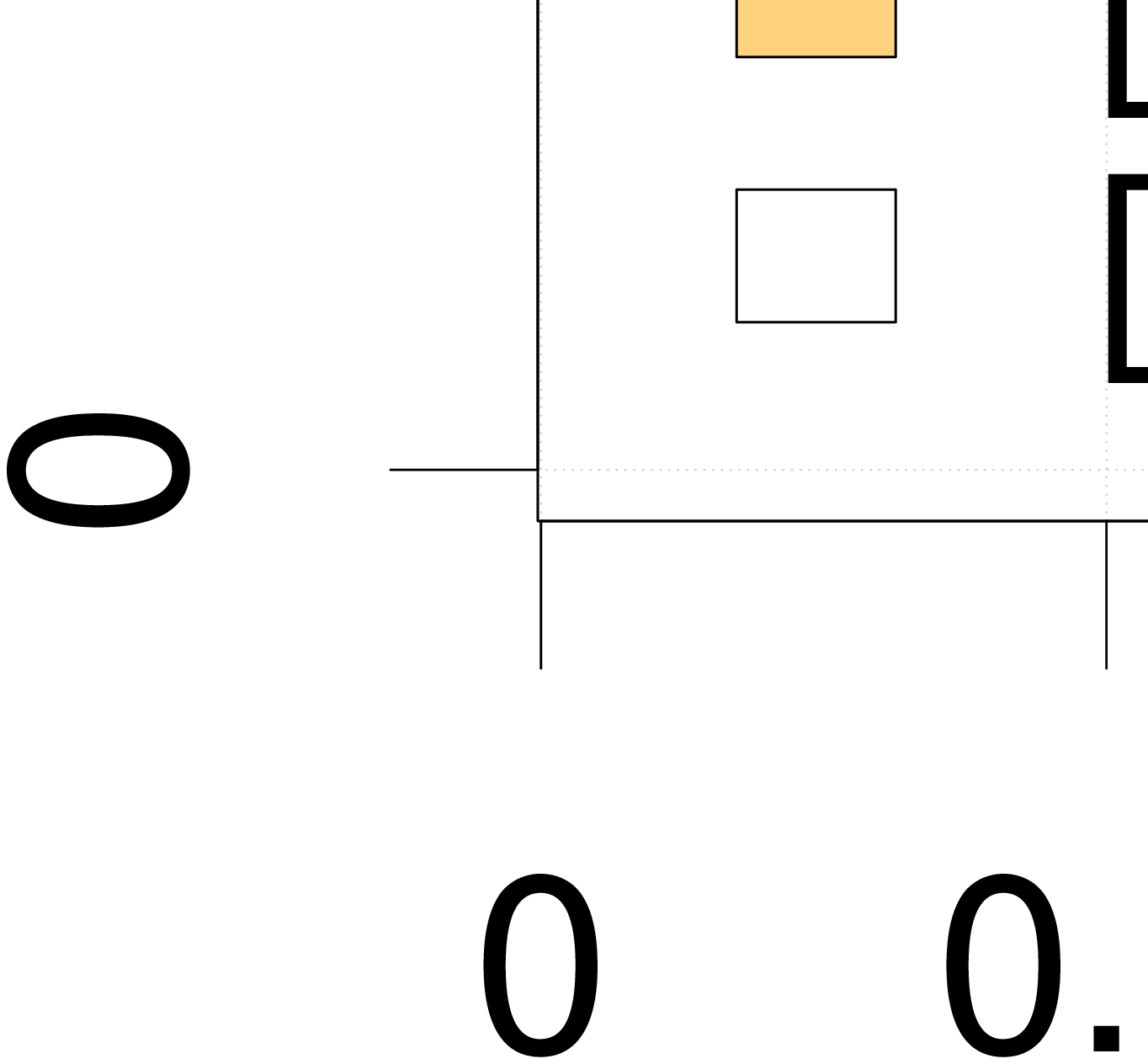}
\caption{Differences between the EAFs of MOEA/D-PS with MOEA/D with small population. The red shade encodes the magnitude of the observed difference on DTLZ1$^{-1}$. The darker the red shade, the better performance of the algorithm. MOEA/D-PS (left) performs well towards all regions of the objective space.}

\label{fig:eaf_inv_DTLZ1_small}
\end{figure}
    % ~~
    % \qquad
\begin{figure}[htbp]
\centering

\includegraphics[width=1\textwidth]{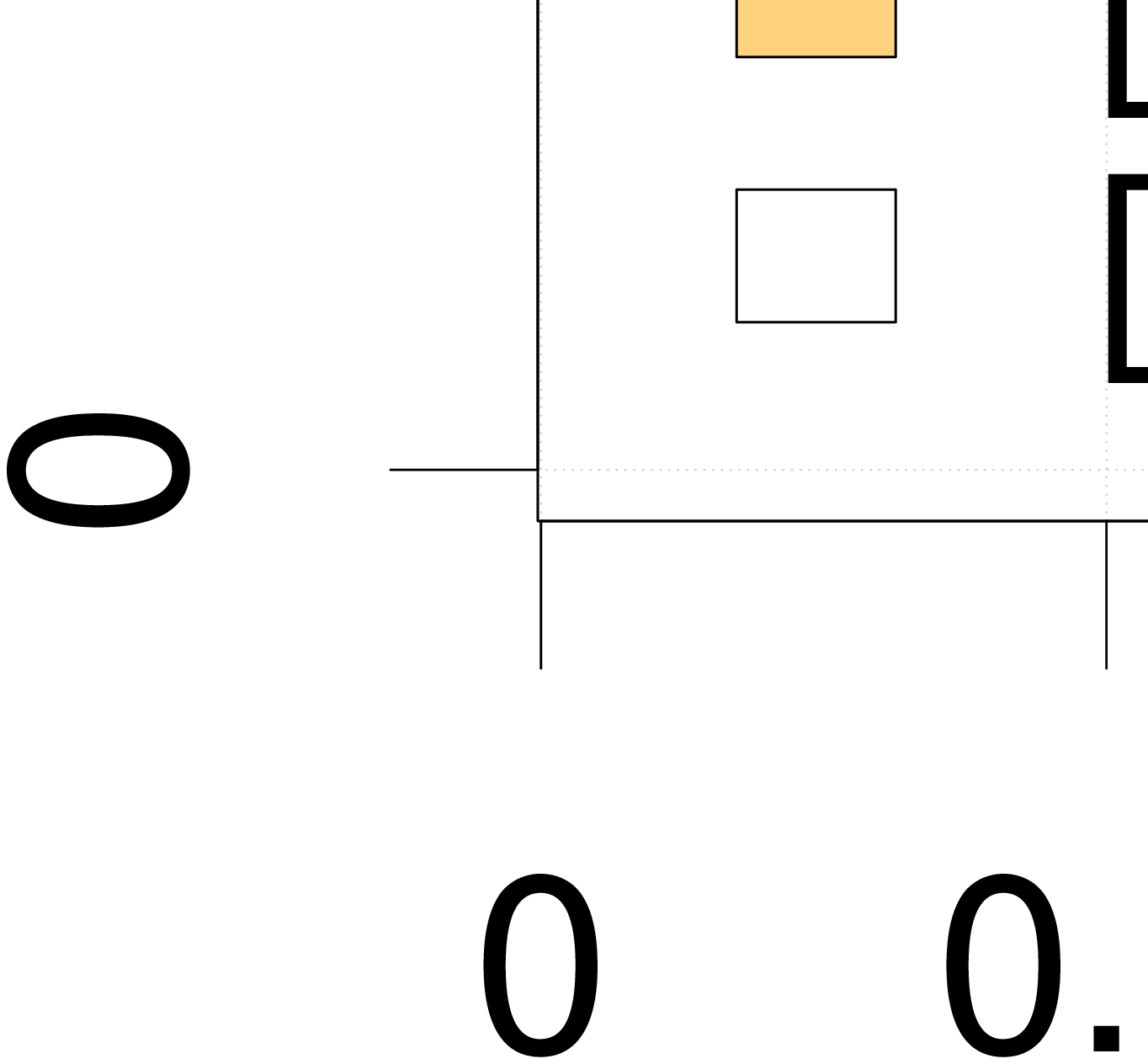}
\caption{Differences between the EAFs of MOEA/D-PS with MOEA/D with big population. The red shade encodes the magnitude of the observed difference on DTLZ1$^{-1}$. The darker the red shade, the better performance of the algorithm. MOEA/D-PS (left) performs better at all regions of the objective space.}

	\label{fig:eaf_inv_DTLZ1_big}
\end{figure}

Figures \ref{fig:anytime_hv_uf10},~\ref{fig:anytime_hv_inv_DTLZ3},~\ref{fig:anytime_hv_DTLZ2},~\ref{fig:anytime_hv_F9} illustrate the anytime performance of the three different MOEA/D variants hypervolume (higher is better) on the UF10, DTLZ3$^{-1}$, DTLZ2 and F9, respectively. We selected those Figures because they represent well the trend behaviour of the algorithms studied in this work. These results indicate that MOEA/D-PS shows an improvement over MOEA/D with a big population, achieving higher HV values in earlier stages of the search. Surprisingly, MOEA/D-PS converges as fast as MOEA/D with a small population.  This similar convergence behaviour is consistent when considering only the working population since the working population dynamics in MOEA/D-PS has the size to the working population dynamics of MOEA/D with small population.

The anytime performance Figures \ref{fig:anytime_hv_uf10},~\ref{fig:anytime_hv_inv_DTLZ3},~\ref{fig:anytime_hv_DTLZ2},~\ref{fig:anytime_hv_F9} together with Table~\ref{chap5:stats_bibbob_hv} reveal the most striking observation to emerge from our conceptual comparison. What stands out is that MOEA/D-PS improves the convergence speed substantially when compared to MOEA/D using the same population size (MOEA/D-PS vs MOEA/D with big population). At the same time, it achieves an impressive higher number of non-dominated solutions (Table~\ref{chap5:stats_bibbob_hv}). Moreover, this observation is not limited to a specific group of MOPs but present in most MOPs studied here, suggesting the stability of MOEA/D with the Partial Update Strategy. This combination of findings provides an explanation of the relationship between MOEA/D-PS with the consistent improvements in performance over the MOEA/D framework.

\subsection{Empirical Attainment Performance}

Figures \ref{fig:eaf_F22_small}, \ref{fig:eaf_F22_big}, \ref{fig:eaf_UF6_small}, \ref{fig:eaf_UF6_big},  \ref{fig:eaf_inv_DTLZ1_small} and \ref{fig:eaf_inv_DTLZ1_big} depict the differences between the EAFs of MOEA/D-PS and the EAF of the MOEA/D with different population sizes, on F22 and DTLZ1$^{-1}$, respectively, in shades of red. \textit{The shades of the red show the amount of the differences of the probabilistic distribution of the outcomes obtained by the algorithms: shades closer to red indicate higher differences between the probability distributions, shades closer to orange indicate little difference, and shades closer to white indicate no difference.} We chose these problems because they represent well the behaviour of the algorithms we study here. The lower line shows the global best set of solutions attained overall runs of all algorithms (grand best attainment surface) in all Figures. In contrast, the upper line shows solutions that are always dominated (grand worst attainment surface). For all Figures, the left panel show regions where the EAF of MOEA/D-PS performs better or worse when compared to the other MOEA/D variants. For Figures \ref{fig:eaf_F22_small}, \ref{fig:eaf_UF6_small} and \ref{fig:eaf_inv_DTLZ1_small}, the right panels show differences in support of MOEA/D with small population size and the right panels on Figures \ref{fig:eaf_F22_big}, \ref{fig:eaf_UF6_big} and \ref{fig:eaf_inv_DTLZ1_big} show the differences in support of MOEA/D with big population size..

Figures~\ref{fig:eaf_F22_small} and \ref{fig:eaf_F22_big} show the EAF for the F22 MOP. We can see that MOEA/D-PS performs better than the other two variations on the centre side of the Figure, attaining this region at least 60\% of the runs, indicated by the shades of red. This result is a rather surprising one. That is because a closer inspection of the Figures shows that MOEA/D-PS has a strong tendency to focus on well-balanced solutions, i.e., solutions that have a good trade-off among the objectives. Figures \ref{fig:eaf_UF6_small} and \ref{fig:eaf_UF6_big} show the same behaviour for the UF6 MOP. However, this focus on central regions only is not as strong for the MOP with a linear-shaped Pareto Front (DTLZ1$^{-1}$, discussed next). 

As above, Figure~\ref{fig:eaf_inv_DTLZ1_small} and \ref{fig:eaf_inv_DTLZ1_big} show, in shades of red, the attainment regions but now for the DTLZ1$^{-1}$. Looking at both Figures, we can see that MOEA/D-PS performs much better than MOEA/D with both small and big population. These big differences suggest that MOEA/D-PS can perform much better in MOPs with a linear-shaped Pareto Front.

\subsection{Statistical Analysis}

\begin{figure}[htbp]
    \centering
    \includegraphics[width=1\textwidth]{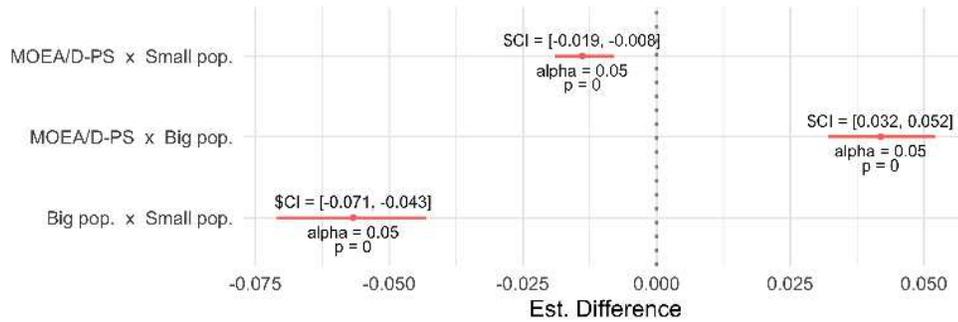}
    \caption{Statistical comparisons between the methods tested using 95\% confidence intervals for the median of paired differences of hypervolume values at 5000 function evaluations between MOEA/D-PS and MOEA/D with big population and MOEA/D with small population. Red intervals with a diamond indicate statistically significant results. MOEA/D-PS has a statistical advantage over MOEA/D with big population and a disadvantage over MOEA/D with small population. This result confirms that MOEA/D-PS increases the convergence speed over MOEA/D with big population, when having a small working population may help MOEA/D-PS find increments in hypervolume convergence speed.}

    \label{fig:conf_int_5000}
\end{figure}
\begin{table}[htbp]
	\centering
	\small
	\caption{Anytime Statistical significance of differences in median HV, for the three algorithms tested in this section, at 5000 function evaluations. Values are Hommel-adjusted p-values of Wilcoxon Rank-sum tests. ``$\uparrow$" indicates superiority of the column method and ``$\leftarrow$'' indicates superiority of the row method ($95\%$ confidence level).}
	\label{chap5:pvals0}

\begin{tabular}{l|ll}
		\hline
		\rowcolor[gray]{.8}\multicolumn{3}{c}{\textbf{5000 function evaluations}}\\\hline
		\rowcolor[gray]{.95}\textbf{} & MOEA/D-PS & Big pop. \\ \hline
		Big pop.  & 2.890526e-11  $\uparrow$   &                    \\ \hline
		Small pop.  & 1.088418e-10  $\leftarrow$   &  9.243532e-13 $\leftarrow$   \\ 

\hline
\end{tabular}
\end{table}

\begin{figure}[htbp]
    \centering
    \includegraphics[width=1\textwidth]{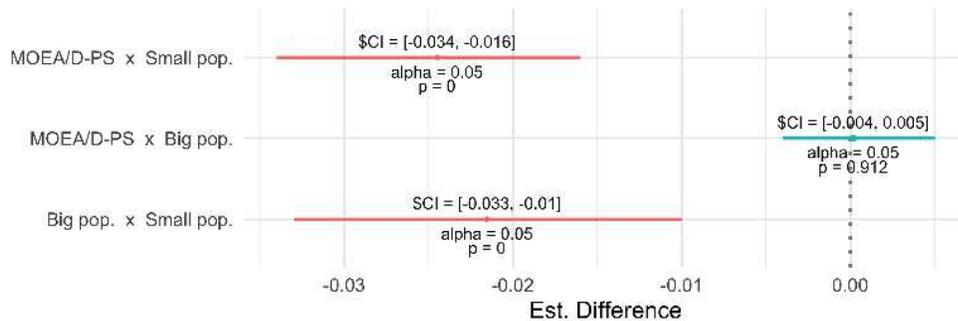}
    \caption{Statistical comparisons between the methods tested using 95\% confidence intervals for the median of paired differences of hypervolume values at 15000 function evaluations between MOEA/D-PS and MOEA/D with big population and MOEA/D with small population. Red intervals with a diamond indicate statistically significant results, and blue intervals with a circle indicate no statistically significant results. MOEA/D-PS has no statistically significant differences to MOEA/D with big population and a disadvantage over MOEA/D with small population. We consider this to be the point in search where MOEA/D with big population start to recover from its slower convergence speed.}
    \label{fig:conf_int_15000}
\end{figure}
\begin{table}[htbp]
	\centering
	\small
	\caption{Anytime Statistical significance of differences in median HV, for the three algorithms tested in this section, at 15000 function evaluations. Values are Hommel-adjusted p-values of Wilcoxon Rank-sum tests. ``$\uparrow$" indicates superiority of the column method, ``$\leftarrow$'' indicates superiority of the row method and ``$\approx$" indicates differences not statistically significant ($95\%$ confidence level).}
	\label{chap5:pvals2}

\begin{tabular}{l|ll}
		\hline
		\rowcolor[gray]{.8}\multicolumn{3}{c}{\textbf{15000 function evaluations}}\\\hline
		\rowcolor[gray]{.95}\textbf{} & MOEA/D-PS & Big pop. \\ \hline
		Big pop.  & 9.124481e-01 $\approx$   &                    \\ \hline
		Small pop.  & 1.169062e-11 $\leftarrow$   & 4.004438e-07 $\leftarrow$   \\ 
\hline
\end{tabular}
\end{table}

\begin{figure}[htbp]
    \centering
     \includegraphics[width=1\textwidth]{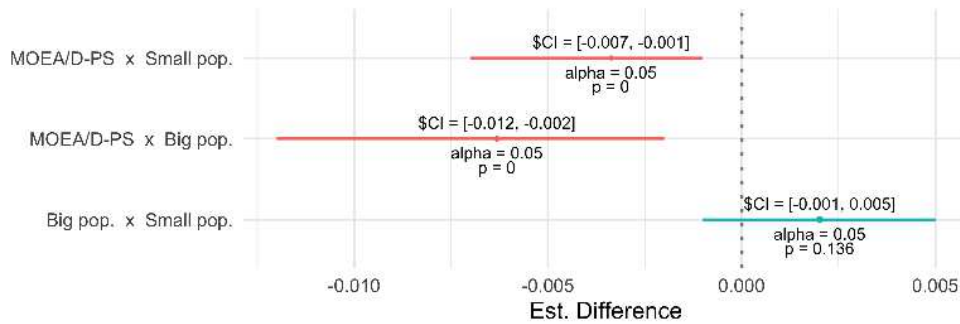}
    \caption{Statistical comparisons between the methods tested using 95\%  confidence intervals on the paired differences between the algorithms, given hypervolume values at 100000 function evaluations. Red intervals with a diamond indicate statistically significant results, and blue intervals with a circle indicate no statistically significant results. MOEA/D-PS has no statistically significant differences to MOEA/D with big population and a disadvantage over MOEA/D with small population. Although all algorithms achieve high hypervolume results in most MOP, we found statistical differences among them at the end of the search. A big population may help MOEA/D find increments in hypervolume.}
    \label{fig:conf_int_100000}
\end{figure}

\begin{table}[htbp]
	\centering
	\small
	\caption{Anytime Statistical significance of differences in median HV, for the three algorithms tested in this section, at 100000 function evaluations. Values are Hommel-adjusted p-values of Wilcoxon Rank-sum tests. ``$\uparrow$" indicates superiority of the column method, ``$\leftarrow$'' indicates superiority of the row method and ``$\approx$" indicates differences not statistically significant ($95\%$ confidence level).}
	\label{chap5:pvals3}

\begin{tabular}{l|ll}
		\hline
		\rowcolor[gray]{.8}\multicolumn{3}{c}{\textbf{100000 function evaluations}}\\\hline
		\rowcolor[gray]{.95}\textbf{} & MOEA/D-PS & Big pop. \\ \hline
		Big pop.  & 1.99 e-04 $\uparrow$   &                    \\ \hline
		Small pop.  & 7.26 e-05 $\uparrow$   & 0.14 $\approx$   \\ 

\hline
\end{tabular}
\end{table}

The current study found that MOEA/D-PS improves the convergence speed substantially when compared to MOEA/D using the same population size (MOEA/D-PS vs MOEA/D with big population, Subsection \ref{chap4:anytime}). Here we want to verify the convergence improvements of MOEA/D-PS over MOEA/D in terms of higher HV values in earlier stages of the search since all methods achieve approximately the same hypervolume result. For that, we perform the Wilcoxon Rank Sum Tests over 5000 and 15000 evaluations. We also consider the final result for completeness, when the computational budget is met, at 100000 evaluations.

Figure ~\ref{fig:conf_int_5000} and Table \ref{chap5:pvals0} present the statistical comparison between MOEA/D-PS, MOEA/D with small population and MOEA/D with big population at 5000 function evaluations. As we can see, MOEA/D-PS has a statistical advantage over MOEA/D with big population and a disadvantage over MOEA/D with small population, confirming our previous analysis than MOEA/D-PS increases the convergence speed over MOEA/D with big population, lower bounded by MOEA/D with small population. The same trend continues until around 15000 function evaluations are met, as shown in Figure~\ref{fig:conf_int_15000} and Table \ref{chap5:pvals2}) when the statistical results show no difference between MOEA/D-PS and MOEA/D with big population. Then, at 100000 (the maximum number of evaluations), we can see at Figure~\ref{fig:conf_int_100000} and Table \ref{chap5:pvals3} that the statistical tests indicate some disadvantage of MOEA/D-PS against the other methods. In summary, we found a significant difference between all methods over the first initial stages of the search, corroborating the results observed in the anytime performance, Subsection \ref{chap4:anytime}). There are statistical differences in the hypervolume values of the final approximation sets, although all algorithms achieve good results in most of the functions. 

\subsection{Influence of the Population Size} 

Another reason for the lower performance of MOEA/D with a small population size might be that the limits imposed by the size of the working population and consequently on the quality of the external archive. We reason that such a small population size negatively influences the external archive because good solutions found by the algorithm during the run might be excluded from the final population. 

On the other hand, MOEA/D-PS might be able to use the larger population as an extra archive population for a small working population. It might be that, for problems where a large population size is needed to search for better-performing solutions, using the PS strategy is useful because the large population size of MOEA/D-PS is playing the role of an archive population for a small working population size. However, more work is needed to clarify the relationship between the archive population and the large population in MOEA/D-PS and in more general, the large population of MOEA/D with RA.

\section{Conclusion}
\label{section:conclusion}

%% Restating the aims of the study
In this work, we investigate the relationship of partially updating the population in MOEA/D using Resource Allocation, represented by MOEA/D-PS, and population size in practice. This study is motivated by the realisation that the Partial Update Strategy uses its control parameter, $n$, to regulate the proportion of the population selected for variation at any iteration, the working population. This working population usually has a small size, resembling MOEA/D variants with small population size. On the other hand, the whole population, which combines both the working population and the population not selected by the Partial Update Strategy, has a considerable size. This size resembles MOEA/D variants with big population size. We extensively study the conceptual correspondences of MOEA/D with the Partial Update and MOEA/D with small population size and big population size on more than 70 MOPs.

%% Summarising main research findings
We found strong evidence that MOEA/D-PS performance is good independently of the MOP in question since MOEA/D with the Partial Update of the population (MOEA/D-PS) always performs as one of the best algorithms in the MOPs studied here. The anytime analysis of the algorithms shows that MOEA/D-PS has a similar convergence behavior as MOEA/D with a small population. That is, MOEA/D-PS quickly finds its best results while maintaining the benefit of a broader exploration of the search space, a common characteristic of evolutionary algorithms with a big population.  The reason relies on the relationship between different populations and their advantages in solving MOPs. That is, a small population can approach the Pareto Front quickly but might not be able to explore and cover this Pareto Front given its limited size; however, a larger population will likely be better at this task, at a higher cost (\cite{glasmachers2014start}). 

%% Suggesting implications for the field of knowledge
The anytime results found in Section~\ref{section:results} show that MOEA/D-PS provides a simple yet efficient approach to mitigate common problems related to population size choice. Examples of such problems are the likely waste of computation resources induced by the large size of a population or the premature stagnation caused by the small size of the working population (\cite{lin2019multi}). Consequently, the results of this work confirm that MOEA/D-PS is an effective and stable improvement over the MOEA/D framework. Moreover, the anytime analysis complements the results of those earlier studies, such as maintaining a big population size while updating a small subset of solutions as in ~\cite{pruvost2020subproblemselection,lavinas2020moea}. Besides, this study demonstrates the partial update strategy ability to increase the convergence speed in most of the more than 70 MOPs explored in this work, as confirmed by the overall good anytime performance. More work is needed to determine the best strategy for deciding which solutions are selected to be updated. 

In addition to the anytime performance, we further analysed the performance of MOEA/D with the Partial Update Strategy using the Empirical Attainment Function (EAF). The EAF results illustrate the hybrid behaviour of MOEA/D-PS, as we can observe that it can find better EAF regions in all problems tested compared to both MOEA/D with a small or a big population. This analysis also suggests that MOEA/D-PS tends to focus on well-balanced solutions, i.e., solutions with a well-balanced trade-off among objectives.

Moreover, our findings have significant implications for the understanding of how MOEA/D with Resource Allocation works. Together, these findings support that using the partial update of the population as one improvement to the MOEA/D framework shall be preferred. As future works, we want to verify whether MOEA/D-PS would benefit from adapting the $n$ value throughout the search and from studying the algorithm behaviour on constrained MOPs such as in the one proposed by~\cite{andrew2014objectives,MoonOrbitingSatellite2015, kohira2018proposal}. 

% Moreover, our findings have significant implications for the understanding of how MOEA/D with Resource Allocation works. Together, these findings support that using the partial update of the population as one improvement to the MOEA/D framework shall be preferred. As future works, we highlight two main directions to extend this work. The first is to verify whether MOEA/D-PS would benefit from adapting the $n$ value throughout the search, especially when the performance improvements between consecutive iterations are small. The second is to study the algorithm performance and behaviour on constrained MOPs, such as the recent multi-modal CEC'19 problems and the real-world simulation-based MOPs~\cite{MoonOrbitingSatellite2015, kohira2018proposal, andrew2014objectives}. 

\small

\bibliographystyle{apalike}
\bibliography{ecjsample}

\end{document}